\documentclass[10pt,twocolumn,letterpaper]{article}

\usepackage[pagenumbers]{cvpr} %

\usepackage[dvipsnames]{xcolor}

\definecolor{myred}{RGB}{220,50,47} %
\definecolor{mygreen}{RGB}{133,153,0}
\definecolor{commentcolor}{RGB}{133,153,0}

\definecolor{mygreen}{rgb}{0.29, 0.7, 0.48}

\newcommand*{\affmark}[1][*]{\textsuperscript{#1}}

\definecolor{cvprblue}{rgb}{0.21,0.49,0.74}
\usepackage[pagebackref,breaklinks,colorlinks,citecolor=cvprblue]{hyperref}

\def\paperID{4161} %
\def\confName{CVPR}
\def\confYear{2024}

\title{AVID: \underline{A}ny-Length \underline{V}ideo \underline{I}npainting with \underline{D}iffusion Model}
\newcommand{\methodAbbr}{AVID}

\author{Zhixing Zhang\affmark[1]\quad
        Bichen Wu\affmark[2] \quad
        Xiaoyan Wang\affmark[2] \quad
        Yaqiao Luo\affmark[2] \quad
        Luxin Zhang\affmark[2] \quad \\
        Yinan Zhao \affmark[2] \quad 
        Peter Vajda\affmark[2] \quad
        Dimitris Metaxas\affmark[1] \quad
        Licheng Yu\affmark[2]\\
        {\affmark[1]Rutgers University\quad\quad\quad\quad\quad\affmark[2]GenAI, Meta }
        }

\begin{document}

\twocolumn[{
\renewcommand\twocolumn[1][]{#1}
\maketitle
\begin{center}
    \centering
    \vspace*{-.5em}
    \includegraphics[width=1\linewidth]{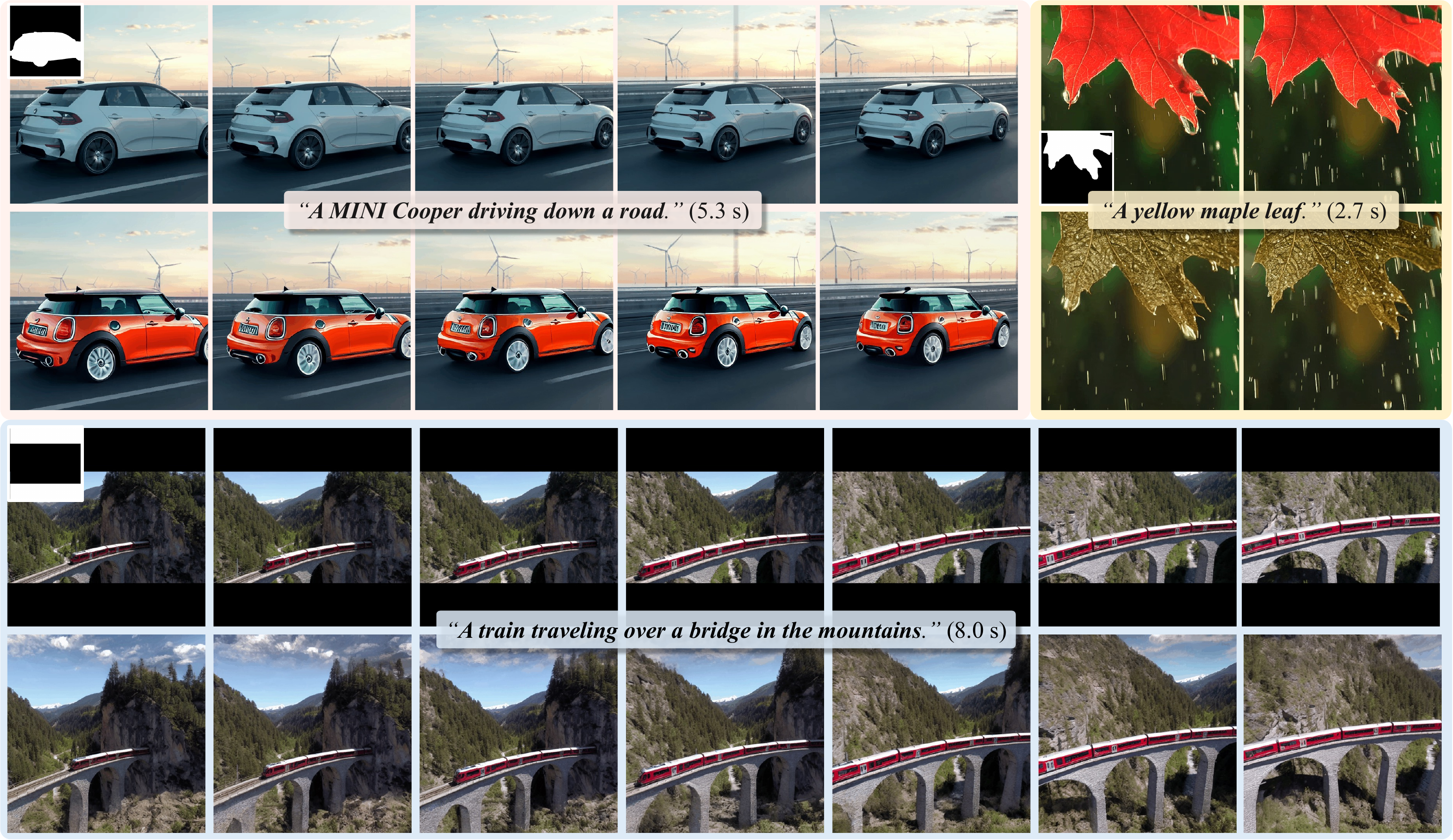}
    \captionof{figure}{
    \textbf{Video inpainting.} 
    We introduce a video inpainting method versatile across a spectrum of video durations and tasks. 
    Displayed frames are uniformly selected from videos of different lengths.
    The first row in the figure contains the source videos and the target regions, while the bottom row shows the results. 
    The caption in the middle represents the language guidance and duration for each video.
    }
    \label{fig: teaser}
\end{center}
}]

\begin{abstract}

Recent advances in diffusion models have successfully enabled text-guided image inpainting.
While it seems straightforward to extend such editing capability into the video domain, there have been fewer works regarding text-guided video inpainting.
Given a video, a masked region at its initial frame, and an editing prompt, it requires a model to do infilling at each frame following the editing guidance while keeping the out-of-mask region intact.
There are three main challenges in text-guided video inpainting: ($i$) temporal consistency of the edited video, ($ii$) supporting different inpainting types at different structural fidelity levels, and ($iii$) dealing with variable video length.
To address these challenges, we introduce Any-Length Video Inpainting with Diffusion Model, dubbed as AVID.
At its core, our model is equipped with effective motion modules and adjustable structure guidance, for fixed-length video inpainting. 
Building on top of that, we propose a novel Temporal MultiDiffusion sampling pipeline with a middle-frame attention guidance mechanism, facilitating the generation of videos with any desired duration. 
Our comprehensive experiments show our model can robustly deal with various inpainting types at different video duration ranges, with high quality\footnote{More visualization results are made publicly available \href{https://zhang-zx.github.io/AVID/}{here}.}.
\end{abstract}

\section{Introduction}
\label{sec: intro}

Past years have witnessed remarkable advancements in image inpainting~\cite{lugmayr2022repaint, xie2023smartbrush, cheng2022inout}, which infills a given masked region in an image.
Owing to the recent explosive evolution of text-to-image (T2I) generative models, the content infilling becomes even more flexible by conditioning on a given textual description~\cite{xie2023smartbrush, rombach2022high}.
On the other hand, text-to-video (T2V) generation is also developing rapidly~\cite{hong2022cogvideo}.
This leads to a compelling inquiry: can we also harness this prowess for text-guided video inpainting?

Imagine being given a video of ``a car traversing a road" and one wants to replace this car with a MINI Cooper.
Perhaps one of the easiest editing approaches is the user just simply clicking on the car in the first frame and writing down a short textual prompt \emph{``MINI Cooper"}, then an advanced model congests this information and generates a spatially-seamless and temporally consistent moving MINI Cooper in the scene while not changing any other video parts.
We believe such interaction will become a fundamental and popular video editing approach, thus decide to address this task as its pioneering work.

There are mainly three main challenges in video inpainting.
First, the synthesized content must exhibit temporal consistency. 
If an entire object is rendered, its identity should persist throughout the video. 
For example, if the color of the car is changed to green, it is imperative that the hue remains consistent from start to finish, the car should remain the same shade of green, rather than transitioning from, say, neon green to a darker variant.
Second, there are various editing types in video inpainting.
For example, as in \cref{fig: teaser}, one common inpainting type is object swap (\eg ``replace a car with a MINI Cooper", another type is re-texturing (\eg ``turn the leaf color from red to yellow"), and uncropping (\eg ``fill in the upper and lower region") is also popular.
Different editing types require different levels of structural fidelity to the original video.
For instance, in object swap, if the editing is to transform a masked person into a statue, the structure and motion of the original person indicated by the mask region should be maintained in the output video. 
In comparison, video uncropping entails filling in blank spaces to augment the view, with no guidance signal from the mask region.
Third, an input video can be of variable length, and thus, we expect a good model should robustly handle any video duration.

In light of these challenges, we propose Any-Length Video Inpainting with Diffusion Model, dubbed as \methodAbbr, a unified framework tailored for video inpainting.
Built upon a text-guided image inpainting framework, we integrate motion modules to ensure temporal coherence within the edited region.
A structure guidance module is also incorporated, adaptable to the varying structural fidelity demands of different video inpainting tasks.
Our innovation also includes a zero-shot generation pipeline, reinforced by a middle-frame attention guidance mechanism, enabling the handling of videos across varying durations.
With the proposed model, given a textual description for the desired modification, along with a mask on the initial frame of a video indicating the area to edit as in \cref{fig: teaser}, we can adeptly modify the content of \emph{variable} video length, aligning them with the desired narrative while conserving intricate details, maintaining temporal consistency, and leaving the region outside the mask unaltered. 

To encapsulate our contributions:
\textbf{(I)} We integrate motion modules~\cite{guo2023animatediff} into a text-to-image inpainting model and optimize it on video sequences, thereby ensuring temporal consistency. 
\textbf{(II)} We propose a structure guidance module tailored for different sub-tasks, so that users can control different degrees of output's structural fidelity towards the input video depending on the task and editing need. 
\textbf{(III)} We incorporate a pioneering zero-shot generation technique, proficiently handling different video lengths without additional training. Concurrently, we introduce a middle-frame attention guidance methodology, ensuring temporal consistency even in elongated video sequences.
\textbf{(IV)} Rigorous evaluation on various inpainting tasks, region areas, and video length demonstrates the robustness of our model.

\section{Related Work}
\label{sec: related}

The success of diffusion models~\cite{ho2020denoising, song2020denoising, song2020score} have enabled advanced image generation~\cite{ramesh2022hierarchical, saharia2022photorealistic, han2021dual, nichol2021glide, ramesh2021zero}. 
These models leverage large-scale text-image datasets~\cite{schuhmann2021laion} to produce remarkable outcomes. 
Further, recent works utilized the pre-trained text-to-image models for image manipulation in response to natural language descriptions~\cite{hertz2022prompt, gal2022image, ruiz2022dreambooth, kawar2022imagic, liu2021more, kwon2022diffusion, liu2022name, miyake2023negative, han2020robust, han2024proxedit, liu2024lepard, liu2023deformer, zhang2023sine, han2023svdiff, song2022diffusion, bar2023multidiffusion, stathopoulos2024score, brooks2023instructpix2pix, kumari2023multi}. 
Among them, customized image generation has particularly benefited from controlling the generated content via additional structure modules, \eg, ControlNet~\cite{zhang2023adding}, T2I-adapter~\cite{mou2023t2i}, \etc.

Such techniques are also influencing a traditional but popular image editing task -- image inpainting.
While generative adversarial networks~\cite{yu2018generative, yu2019free} have been mainly applied for this task, recent diffusion-based models~\cite{saharia2022palette, anciukevivcius2023renderdiffusion} are showing more impressive results. 
However, these models are limited in filling in the content using out-of-mask context only. 
A more flexible usage is adding text control~\cite{yang2023uni, yang2023paint, avrahami2023blended, avrahami2022blended, wang2023imagen, xie2023smartbrush} that allow for text-guided image inpainting.
Latent Blended Diffusion~\cite{avrahami2023blended} proposed blending the generated and original image latents, Imagenator~\cite{wang2023imagen} and Diffusion-based Inpainting~\cite{rombach2022high} fine-tune pre-trained text-to-image generation models with masked images as additional input, and SmartBrush~\cite{xie2023smartbrush} fine-tunes an additional mask prediction branch on object-centric datasets.

Extending the success of text-guided image inpainting to video domain presents unique challenges, especially in maintaining the temporal consistency in videos of arbitrary duration. 
The scarcity of high-quality, large-scale video datasets~\cite{hong2022cogvideo, villegas2022phenaki, wu2021godiva, wu2022nuwa, singer2022make, han2022show, ho2022video, blattmann2023align, ho2022imagen, guo2023animatediff, chen2023videocrafter1} further complicates this task. 
Recent efforts have explored utilizing pre-trained image models for video editing, \eg employing DDIM inversion~\cite{song2020denoising} for consistent latents~\cite{shin2023edit, wu2023tune, qi2023fatezero, ceylan2023pix2video, wang2023zero, geyer2023tokenflow}. 
Nevertheless, most approaches are proposed without considering an explicit mask input. 
Relying on the textual editing prompts only could easily change the undesired regions.
As comparison, VideoComposer~\cite{wang2023videocomposer} takes the masked frames as input and addresses video inpainting as one of their editing tasks, yet its constraint of applying a uniform target region across all frames limits its flexibility and sacrifices the editing quality.

In this work, we introduce a simple and effective framework for text-guided video inpainting. 
Our approach incorporates motion modules into a pre-trained text-to-image diffusion model, ensuring ts temporal coherence, and integrates a structure guidance module to satisfy varying structural fidelity needs. 
Moreover, we present an inference pipeline capable of handling videos of any duration, paving the way for practical applications in real-world scenarios.

 \section{Methods}
\label{sec: methods}

\begin{figure*}
\setlength{\linewidth}{\textwidth}
\setlength{\hsize}{\textwidth}
\centering
\includegraphics[width=1\linewidth]{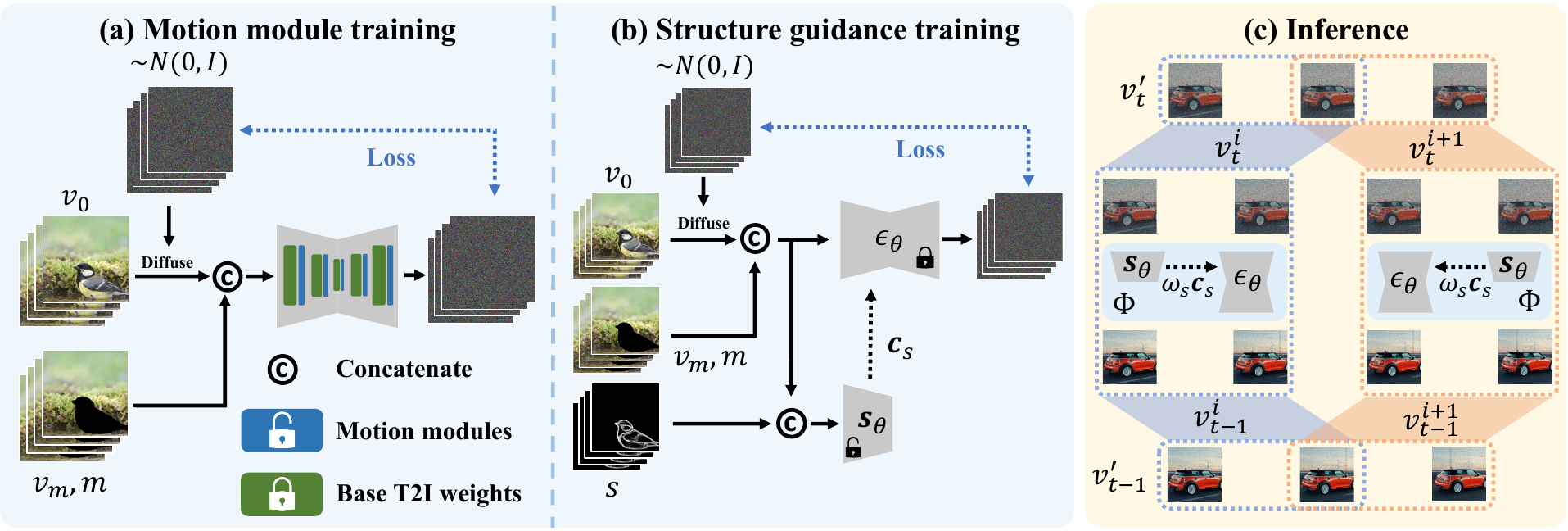}
\caption{\textbf{Overview of our method.}
In the training phase of our methodology, we employ a two-step approach. 
(a) Motion modules are integrated after each layer of the primary Text-to-Image (T2I) inpainting model, optimized for the video in-painting task via synthetic masks applied to the video data. 
(b) During the second training step, we fix the parameters in the UNet, $\epsilon_\theta$, and train a structure guidance module $\mathbf{s}_\theta$, leveraging a parameter copy from the UNet encoder.
During inference, (c), for a video of length $N^\prime$, we construct a series of segments, each comprising $N$ successive frames.
Throughout each denoising step, results for every segment are computed and aggregated. 
}
\label{fig: overview}

\vspace{-1em}
\end{figure*}

For a video of arbitrary duration, a mask region, and an editing prompt, our objective is to fill in the indicated region at each frame following the editing guidance, while keeping the out-of-mask video portion unchanged.
We introduce our method, \methodAbbr, as depicted in \cref{fig: overview}.
Our approach is built on top of a diffusion-based text-guided image inpainting model~\cite{rombach2022high},
then adapt it to video inpainting by inflating the base model and learning the motion modules (\cref{sec: methods: t2v inpainting}). 
We further add a structure guidance module to meet different inpainting types and structural fidelity requirements (\cref{sec: methods: structure}).
Additionally, we propose a zero-shot inference pipeline to deal with videos of varying lengths (\cref{sec: methods: inference}).

\subsection{Preliminaries} \label{sec: methods: preliminaries}
The diffusion model is defined to approximate the probability density of training data by reversing the Markovian Gaussian diffusion processes~\cite{ho2020denoising}.
Consider an input image $x_0$, we conduct a forward Markov process described as:
\begin{equation}
\small
\label{equ: diffuse}
q\left(x_t \mid x_{t-1}\right)=\mathcal{N}\left(\sqrt{1-\beta_t} x_{t-1}, \beta_t \mathrm{I}\right),
\end{equation}
where $t = 1, \ldots, T$ indicates the number of diffusion steps, with $\beta_t$ controlling the noise level at each step.
A neural network $\epsilon_\theta$ learns to reverse this process, approximating noise $\epsilon_t$ to restore $x_{t-1}$ from $x_t$ using the relation $x_{t-1} = \frac{1}{\sqrt{\alpha_t}}\left(x_t-\frac{1-\alpha_t}{\sqrt{1-\bar{\alpha}_t} } \epsilon_t \right)$, with $\alpha_t = 1 - \beta_t$ and $\bar{\alpha}_t=\prod_{i=1}^t \alpha_i$, as per~\cite{ho2020denoising}.
For conditional diffusion, in our case, text-guided inpainting, we introduce conditions into $\epsilon_\theta$ without altering the process.
Our training objective can be formulated as:
\begin{equation}
    \small
    \label{equ: img loss}
    \mathcal{L} = \mathbb{E}_{\epsilon \sim \mathcal{N}(0, I)}\left[\left\|\epsilon-\epsilon_\theta\left(x_t, t, \mathbf{c}\right)\right\|_2^2\right],
\end{equation}
where $\mathbf{c}$ denotes the conditional inputs.
In our case, $\mathbf{c}=\left(x_m, m, \tau_\theta(y)\right)$, where $m$ is a binary mask indicating the region to modify, $x_m=x_0 \odot (1 - m)$ is the region to preserve, $y$ represents the corresponding textual description, while $\tau_\theta \left( \cdot \right)$ embodies a text encoder that transposes the string into a sequence of vectors. Efficient sampling approaches, such as DDIM~\cite{song2020denoising} or PNDM~\cite{liu2022pseudo}, and classifier-free guidance~\cite{ho2022classifier} can be applied during inference.

\subsection{Text-guided Video Inpainting} \label{sec: methods: t2v inpainting}
Given a video $v_0 = \{x_0^i\}_{i=1}^{N}$ and a mask region at its first frame $m^0$, we seek to edit the target region at each frame, aligning its content with the given text prompt $y$.
In order to predict a precise editing region at each frame, we first propagate $m^0$ to every following frame and obtain a mask sequence $m=\{m^i\}_{i=1}^{N}$.
This is achieved by applying XMem ~\cite{cheng2022xmem} to track the masked region through the whole video, which we found simple and precise for editing types like object swap and re-texturing.
For uncropping, we simply use the same to-fill region for all frames.

To pursue good temporal consistency across video frames, we follow AnimateDiff~\cite{guo2023animatediff} to inflate an image diffusion model by converting its 2D layers to pseudo-3D and adding additional motion modules to learn the temporal correlations between frames.
As described in ~\cref{equ: diffuse} to obtain $v_t = \{x_t^i\}_{i=1}^{N}$, the video frames undergo a forward diffusion process individually.
Then our objective for video inpainting becomes:
\begin{equation}
    \small
    \label{equ: vid loss}
    \mathcal{L} = \mathbb{E}_{\epsilon \sim \mathcal{N}(0, I)}\left[\left\|\epsilon-\epsilon_\theta\left(v_t, t, \mathbf{c}\right)\right\|_2^2\right],
\end{equation}
with $\mathbf{c}=\left(v_m, m, \tau_\theta(y)\right)$ and $v_m = \{x_0^i \odot (1 - m ^i)\}_{i=1}^{N}$.
We optimize only the motion modules to retain the generative capabilities of the pre-trained model, as in ~\cref{fig: overview} (a).

\subsection{Structure Guidance in Video Inpainting} \label{sec: methods: structure}
\begin{figure*}[ht]
\setlength{\linewidth}{\textwidth}
\setlength{\hsize}{\textwidth}
\centering
\includegraphics[width=1\linewidth]{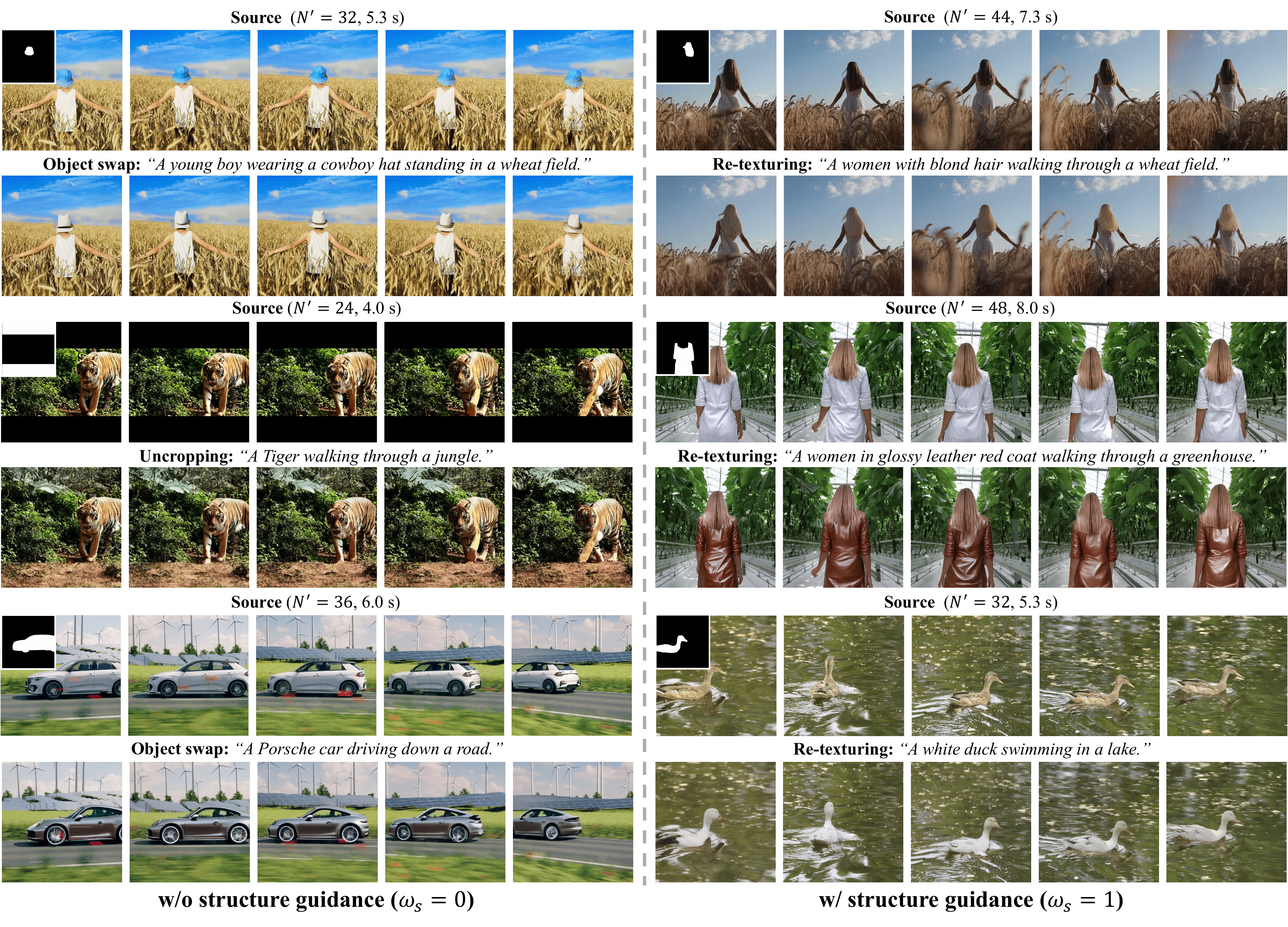}
\vspace{-2em}
\caption{\textbf{Editing on videos of different durations.} We employ our method on various videos and edit them for different tasks. 
We show the wide range of edits our approach can be used with different region sizes and video durations.
Above each video, we note the number of frames, $N^\prime$, and the video duration. 
}
\label{fig: qual}
\vspace{-1.em}
\end{figure*}

In practice, as in ~\cref{fig: teaser}, video inpainting involves different editing types, with respect to different structural fidelity.
For example, re-texturing requires the structure of the source video to be preserved, \eg, converting the material of the person's coat to leather, as in ~\cref{fig: qual}, while uncropping has no such requirement.
This inspires us to introduce an additional structure guidance module to our model.
Following the design of ControlNet~\cite{zhang2023adding}, we fix the parameters of $\epsilon_\theta$ and proceed to train the structure-conditioned module, denoted as $\mathbf{s}_\theta$, as in ~\cref{fig: overview} (b).
We employ a structure information extractor $\mathcal{S}$ to obtain structure condition on each frame $s=\{s^i\}_{i=1}^{N}$, where $s^i = \mathcal{S} \left( x_0^i \right)$.
The structure guidance module outputs $\mathbf{c}_s = \mathbf{s}_\theta\left(v_t, t, \mathbf{c}, s\right)$, composed of $13$ feature maps $\{h_i\}_{i=1}^{13}$ at $4$ different resolutions.
These feature maps are then integrated into the skip connections and the output of the middle block of $\epsilon_\theta$, contributing to the generation process.
Therefore, the training objective for this phase is formulated as:
\begin{equation}
    \small
    \label{equ: stru loss}
    \mathcal{L} = \mathbb{E}_{\epsilon \sim \mathcal{N}(0, I)}\left[\left\|\epsilon-\epsilon_\theta\left(v_t, t, \mathbf{c}, \mathbf{c}_s \right)\right\|_2^2\right],
\end{equation}
with $\mathbf{s}_\theta$ parameters being optimized.

During inference, the structural fidelity is modulated by a scaling factor $\omega_s$ applied to $\mathbf{c}_s$, with the formulation $\epsilon_\theta\left(v_t, t, \mathbf{c}, \mathbf{c}_s \cdot \omega_s\right)$.
Here, $\mathbf{c}_s \cdot \omega_s = \{h_i \cdot \omega_s \}_{i=1}^{13}$, where the scaling factor is applied to each feature map individually.
A higher $\omega_s$ leads to better structural fidelity.

\subsection{Zero-shot Inference for Long Videos} \label{sec: methods: inference}
Although motion modules (implemented as temporal self-attention layers) can take videos with varying numbers of frames theoretically, it suffers from drastic quality degradation when generating a longer video of more frames than training~\cite{guo2023animatediff}.
We show such quality degradation in \emph{supplementary material}.
Inspired by MultiDiffusion~\cite{bar2023multidiffusion} generating a high-resolution image seamlessly consisting of multiple patches, we extend such idea into the temporal axis, making it work effectively to deal with a video of any length.
We further propose a novel middle-frame attention guidance mechanism, to keep the identity unchanged throughout the video.

\noindent \textbf{Temporal MultiDiffusion.} 
We denote our model learned from ~\cref{sec: methods: t2v inpainting} and ~\cref{sec: methods: structure} as $\Phi_{\{\epsilon_\theta, \mathbf{s}_\theta\}}$, such that $v_{t - 1} = \Phi\left(v_t\right)$.
Given a video with $N^\prime$ frames longer than training video length $N$, 
we first segment such longer video $v_t^\prime$ into overlapping clips, 
achieved by applying a sliding window of $N$ frames with a stride of $\mathbf{o}$.
This process partitions $v_t^\prime$ into a series of clips $\{v_t^i\}_{i=1}^n$, where each $v_t^i$ contains $N$ frames, and $n=\lceil \frac{N^\prime - N}{\mathbf{o}} \rceil + 1$ is the total number of clips.
Applying our model once on each clip, we can get one-step denoising $\{v_{t - 1}^i\}_{i=1}^n$, where $v_{t - 1}^i = \Phi\left(v_t^i\right)$.

For any given frame $v_t^\prime[k]$ within $v_t^\prime$, we identify the set of clip indices $\mathbb{S}_k$ that contain this frame. 
Subsequently, for each index $i$ in $\mathbb{S}_k$, the corresponding frame from $v_t^\prime[k]$ is mapped to the $j$-th frame in $v_t^i$, denoted as $v_t^i[j]$. 
Similar to ~\cite{bar2023multidiffusion}, we determine the frame $v_{t-1}^\prime [k]$ in the output video by averaging the results from all relevant clips:
\begin{equation}
\label{equ: multidiff}
\small
v_{t-1}^\prime [k] = \frac{1}{\lVert \mathbb{S}_k \rVert} \sum_{i \in \mathbb{S}_k} v_{t - 1}^i[j],
\end{equation}
where $v_{t - 1}^i[j]$ is the processed frame from $v_t^i$ corresponding to $v_t^\prime[k]$, and  $\lVert \mathbb{S}_k \rVert$  is the cardinality of the set $\mathbb{S}_k$, indicating the number of times frame $k$ is processed across all clips.

\begin{figure}
  \centering
  \includegraphics[width=1\linewidth]{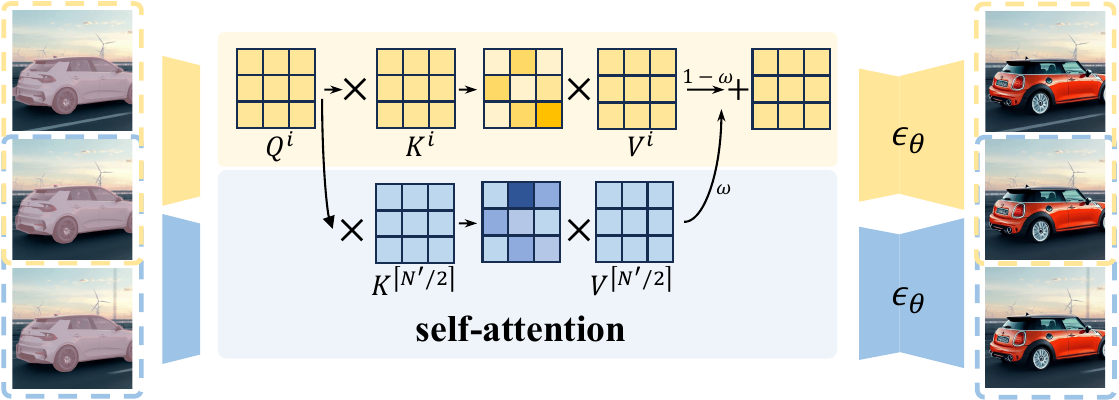}
  \caption{\textbf{Middle-frame attention guidance.} 
  At inference, during each denoising step and within every self-attention layer, we retain the $K^{\lceil N^\prime / 2 \rceil}$ and $V^{\lceil N^\prime / 2 \rceil}$ values from the frame in the middle of the video.
  For the video's $i^{th}$ frame, we utilize its pixel queries, denoted as $Q^i$, to compute an auxiliary attention feature map.
  This is subsequently fused with the existing self-attention feature map within the same layer. 
  }
  \label{fig: attention}

  \vspace{-1.em}
\end{figure}

\noindent \textbf{Middle-frame attention guidance.}
With Temporal MultiDiffusion, we show the smoothness of our long video can be hugely improved \cref{fig: ab_multi}.
However, another critical issue arises
-- identity gradually changes from the initial frame to the last, and such issue could become more and more severe as longer video, as shown in \cref{fig: ab_attn}.

To address this, we introduce a novel attention guidance mechanism, enforcing identity consistency across clips. 
As in \cref{sec: methods: t2v inpainting}, we inflate each self-attention layer to pseudo-$3\text{D}$ self-attention layers.
We denote the input to a self-attention layer of a video frame as $\psi^i$ and $W^Q, W^K, W^V$ as the attention weights, then $Q^i= W^Q \psi^i$, $K^i = W^K \psi^i$, and $V^i = W^V \psi^i$. 
For each self-attention layer, we use the features from the middle frame as guidance, $\psi^{\lceil N^\prime / 2 \rceil}$.
We chose the middle frame due to the overall distance between it and the other frames being the smallest. 
Thus different clips can be more easily connected when their attention is regularized to the same reference frame, mitigating the identity shift issue.
A qualitative study of the key frame selection is provided in \emph{supplementary material}.
The attention formulation for each frame is thus:
\begin{equation}
\small
\label{equ: attn}
\begin{split}
    \small
    \operatorname{Attention}(\psi^i) =& \operatorname{softmax}\left(\frac{Q^i {K^i}^T}{\sqrt{d}} \right)V^i \cdot (1 - \omega)+ \\
                                  & \operatorname{softmax}\left(\frac{Q^i {K^{\lceil N^\prime / 2 \rceil}}^T}{\sqrt{d}}  \right )V^{\lceil N^\prime / 2 \rceil} \cdot \omega ,
\end{split}
\end{equation}
where $\omega$ is a parameter controlling the strength of the guidance signal. 
We provide a visual illustration of the attention guidance in ~\cref{fig: attention}.

\section{Experiments}
\label{sec: experiments}

\begin{figure*}
\setlength{\linewidth}{\textwidth}
\setlength{\hsize}{\textwidth}
\centering
\includegraphics[width=1\linewidth]{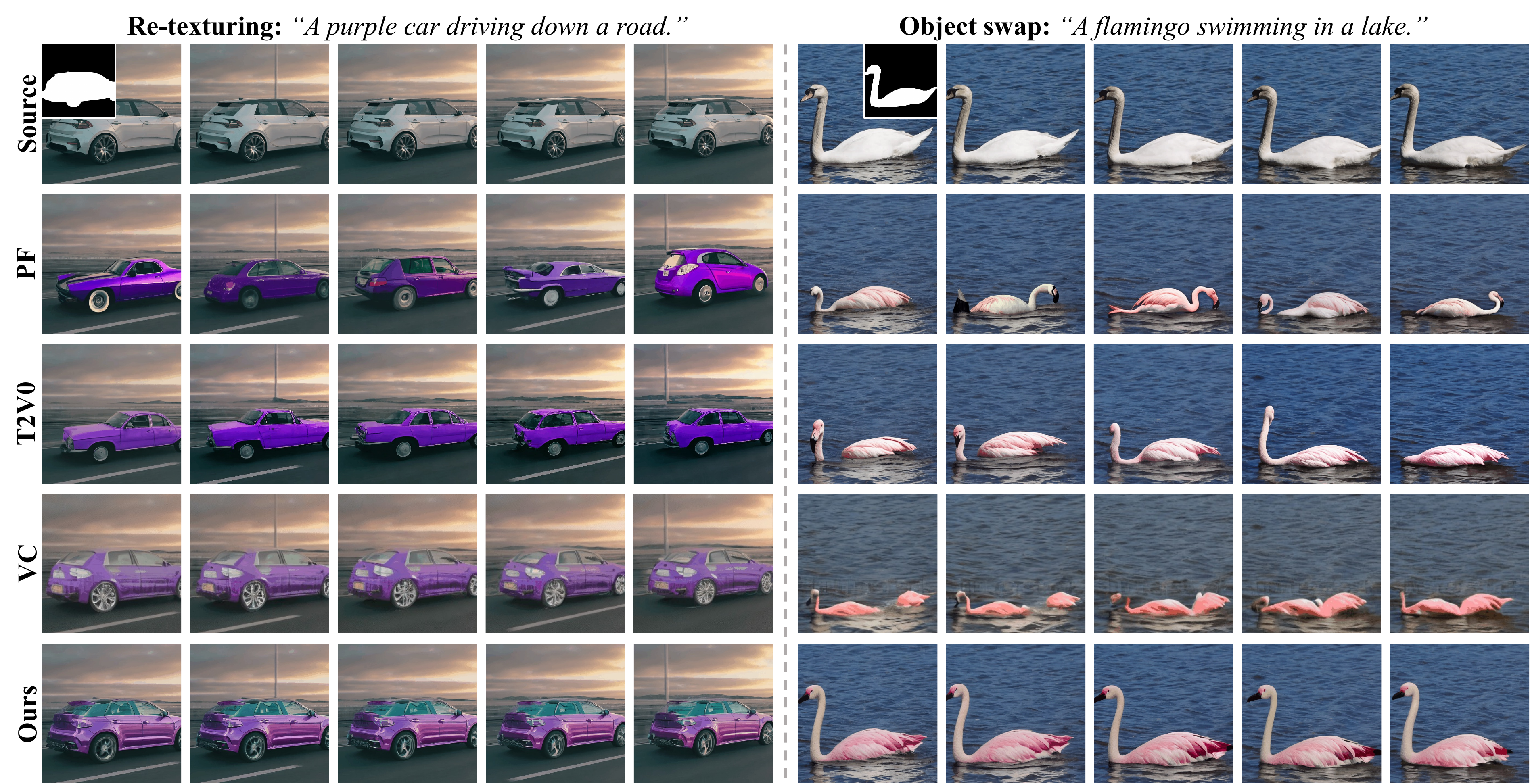}
\caption{\textbf{Comparison of various methods.} 
We compare our method against several approaches, including per-frame inpainting using \emph{text-to-image LDM inpainting} (PF)~\cite{rombach2022high}, \emph{Text2Video-Zero} (T2V0)~\cite{khachatryan2023text2video}, \emph{VideoComposer} (VC)~\cite{wang2023videocomposer}. 
All methods are evaluated using their default hyper-parameters as specified in either their corresponding publications or source codes. 
Each video in our experiments consists of $16$ frames. 
Our proposed approach successfully edits the videos as intended while retaining the details outside the designated target region. 
Moreover, our method upholds the editing capabilities of the image in-painting model we utilized.
Notably, our results demonstrate remarkable consistency, outperforming other methods in our comparison.
}
\label{fig: comp: qual}

\vspace{-1em}
\end{figure*}

\noindent\textbf{Implementation details.}
Our implementation is built upon a large-scale pre-trained inpainting Latent Diffusion Model (LDM)~\cite{rombach2022high}.
For training, we use the watermark-removed Shutterstock video dataset~\cite{Shutterstock}, with motion modules being trained using $16$ frames at a $512\times 512$ resolution using synthetic random mask. 
Subsequently, the parameters from the UNet encoder are transferred to the control module, which is then trained using the same dataset.
We use Holistically-Nested Edge Detection (HED) ~\cite{xie2015holistically} within the synthetic region as the structure guidance for the control module training.
During the structure guidance module training, all parameters in the control module are optimized. 

We evaluate our method on $125$ videos (unseen during training).
Objects designated for editing are pinpointed in these videos. 
On the initial frame of each video, object regions are identified using Grounding-DINO~\cite{liu2023grounding} and Segment-Anything (SAM)~\cite{kirillov2023segment}. 
Subsequent segmentation for the entirety of the video is achieved using XMem~\cite{cheng2022xmem}. 
We further prepare three editing types for evaluation: object swap, re-texturing, and uncropping.
Llama~\cite{touvron2023llama} is then tasked with generating multiple editing prompts for every object within each video, tailored to diverse tasks. 
The FPS for both training and inference are set to $6$.

In~\cref{sec: methods: structure}, we propose to use a scaling factor to flexibly adjust the structural fidelity.
Specifically, object swap and uncropping requires no structural information from the original video, and thus we set the scaling factor for structural fidelity, $\omega_s = 0$.
Conversely, re-texturing, leverages structural data for superior generative outcomes, we empirically set $\omega_s = 1$.
For all long video inference, we set the middle-frame attention guidance scale $\omega$ to $0.3$ and stride for temporal window $\mathbf{o} = 4$ unless otherwise noted.

\subsection{Qualitative Results} \label{sec: experiments: qualitative} 
To comprehensively evaluate the capabilities of our method, we test it on videos of various durations across different inpainting types. 
As shown in ~\cref{fig: qual}, our zero-shot inference approach as in~\cref{sec: methods: inference} is capable of performing diverse editing types, catering to a wide range of mask sizes, and dealing with variable video durations. 
Our method adeptly modifies the specified region without affecting the surrounding content. 
Additionally, it keeps the identity (color, structure, \etc) of the generated content consistent across video frames.

\subsection{Comparisons} \label{sec: experiments: comp}

\begingroup
\setlength{\tabcolsep}{3.1pt} %
\renewcommand{\arraystretch}{0.9} %
\vspace{-.5em}
\begin{table}[!htb]\centering
\small
\begin{tabular}{l ccc ccc ccc}
Task & \multicolumn{3}{c}{Uncropping} & \multicolumn{3}{c}{Object swap} & \multicolumn{3}{c}{Re-texturing$^*$} \\
\toprule
Metric & {\footnotesize BP} & {\footnotesize TA} & {\footnotesize TC} & 
        {\footnotesize BP} & {\footnotesize TA} & {\footnotesize TC} & 
        {\footnotesize BP} & {\footnotesize TA} & {\footnotesize TC} \\
\midrule
PF &43.1 & 31.3 & 93.6 & 41.4 & 31.1& 92.5& 41.4 & 31.2 & 92.4 \\
T2V0 &49.0 & {\bf 31.4} &96.5 &47.3 & 30.1 & 94.9& 47.9 & 30.6& 95.0 \\
VC &55.7 & 31.2&96.4 & 71.0 & {\bf 31.5}& 96.5& 64.5& {\bf 32.1}& 95.5  \\
Ours &{\bf 42.3} & 31.3 & {\bf 97.2}& {\bf 41.1}& 31.5& {\bf 96.5}& {\bf 40.7}& 32.0& {\bf 96.3} \\
    \bottomrule
\end{tabular}
\caption{\textbf{Quantitative results.} We compare our method against several approaches, including per-frame in-painting (PF) using \textit{Stable Diffusion In-painting}~\cite{rombach2022high}, \textit{Text2Video-Zero} (T2V0)~\cite{khachatryan2023text2video}, and \textit{VideoComposer }(VC)~\cite{wang2023videocomposer} on different video inpainting sub-tasks and evaluate generated results using different metrics, including background preservation (BP $\times 10^{-3}$, $\downarrow$ better), text-video alignment (TA, $\uparrow$ better), and temporal consistency (TC, $\uparrow$ better). $^*$ indicates structure guidance is applied for VC and our approach.
}
\label{table: quan}

\vspace{-.5em}

\end{table}
\endgroup

\noindent \textbf{Qualitative comparisons.}
We present a comprehensive evaluation of our method against other diffusion-based video inpainting techniques, notably the per-frame inpainting techniques using inpainting LDM~\cite{rombach2022high} and VideoComposer~\cite{wang2023videocomposer}. 
We apply our model on the same videos as in~\cite{Vats_2023}, which have provided results of VideoComposer~\cite{wang2023videocomposer}.
Recognizing that video inpainting can be construed as a text-to-video generation with set boundary conditions, we also evaluate the training-free Text2Video-Zero~\cite{khachatryan2023text2video} model on top of inpainting LDM as another comparison. 
To highlight the robustness of our foundational model, we limit our evaluations to videos spanning $16$ frames, intentionally omitting the zero-shot long video inferences introduced in ~\cref{sec: methods: inference} for fair comparison.

\cref{fig: comp: qual} (left) compares the performance on re-texturing. 
Note both our model and VideoComposer~\cite{wang2023videocomposer} can apply the structural guidance. 
We let VideoComposer~\cite{wang2023videocomposer} integrate more cues (including the sketch maps, depth maps, and motion vectors) for better results, while our model only applies HED.
\cref{fig: comp: qual} (right) proceeds object swap without structural guidance for all methods.
Specifically, the comparison between VideoComposer~\cite{wang2023videocomposer} (row 4) and our model (row 5) reveals an extraneous color shift in the former, whereas the latter demonstrates impeccable background preservation.
Furthermore, compared with the other two zero-shot methodologies, per-frame editing (row 2) and Text2Video-Zero (row 3)~\cite{khachatryan2023text2video}, our method exhibits far better temporal consistency.
For example, our model can consistently keep the synthesized purple car shape in \cref{fig: comp: qual}, while Text2Video-Zero~\cite{khachatryan2023text2video} presents temporal inconsistencies.

\begin{figure}
    \centering
    \vspace{-.5em}
    \includegraphics[width=.9\linewidth, trim={5em 2em 17em 4em}, clip]{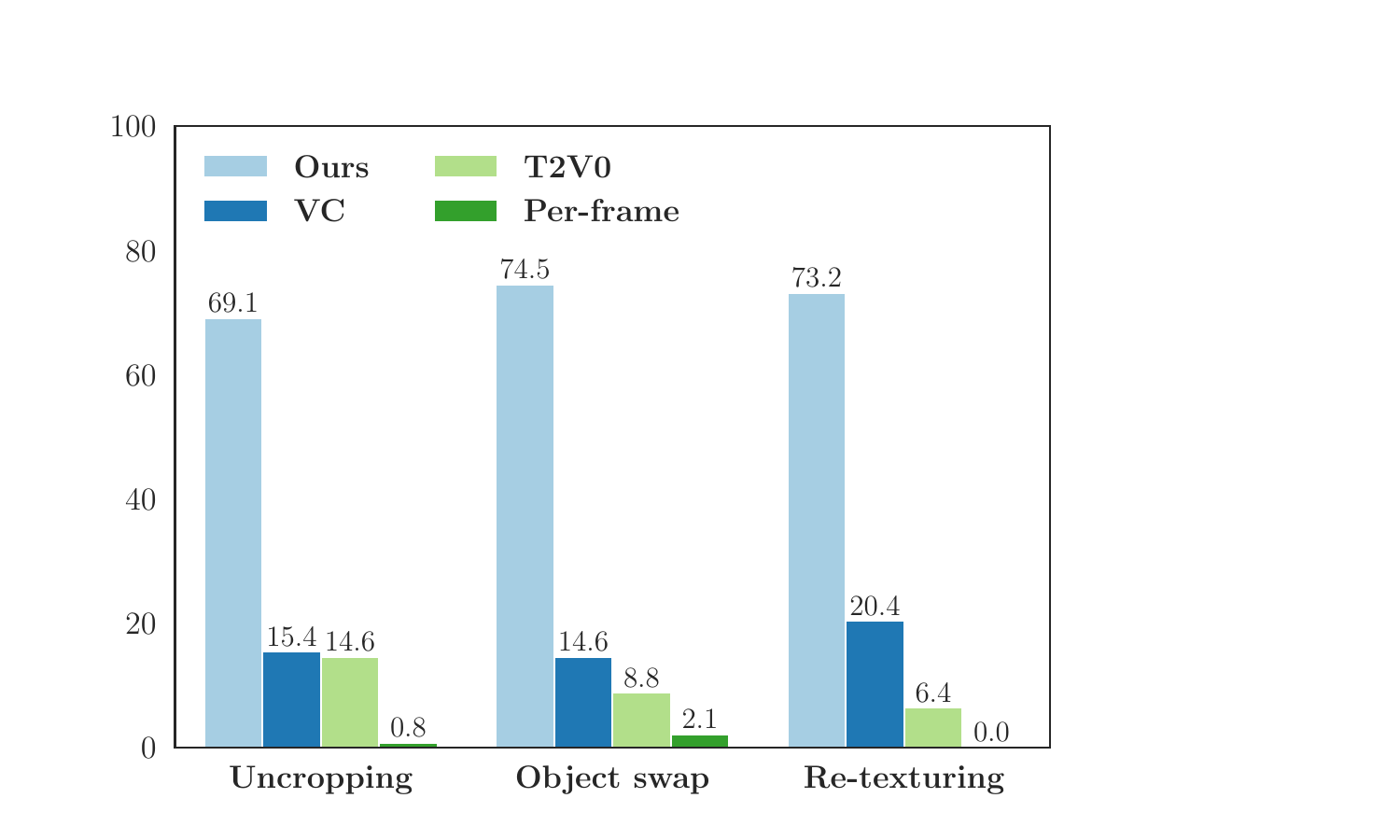}
    
    \caption{\textbf{User study.}
    In our user preference studies, we juxtaposed our method against per-frame in-painting techniques by evaluating prominent models such as \textit{Diffusion-based Image In-painting}\cite{rombach2022high}, \textit{Text2Video-Zero} (T2V0)\cite{khachatryan2023text2video}, and \textit{VideoComposer} (VC)\cite{wang2023videocomposer}, assessing their performances across various tasks. 
    }
    \vspace{-1.5em}
    \label{fig: quan}
    
\end{figure}

\noindent \textbf{Quantitative comparisons.}
\emph{(a) Automatic Metric Evaluation.}
Our model's performance is further quantified using three automatic evaluation metrics. 
Background preservation is measured using the $\text{L}1$ distance between the original and the edited videos within unaltered regions. 
The alignment of the generated video with the text description is evaluated using the CLIP-score~\cite{esser2023structure, hessel2021clipscore}. 
Temporal consistency is assessed by computing the cosine similarity between consecutive frames in the CLIP-Image feature space, as per ~\cite{esser2023structure}. 
As shown in ~\cref{table: quan}, our model exhibits excellent temporal consistency without compromising per-frame quality, as indicated by the text-video alignment scores.
However, we acknowledge that CLIP score may not always correlate with human perception~\cite{molad2023dreamix, wang2023zero}, thus further conducted a user study.
\emph{(b) User study.}
We evaluated our model's effectiveness via user study.
Specifically, the annotators are presented with videos processed on three inpainting types: uncropping, object swap, and re-texturing.
They were asked to judge the overall quality of the video inpainting based on temporal consistency, text-video alignment, and background preservation. 
Results illustrated in \cref{fig: quan} indicate that our model was greatly favored in producing the best outcomes in all types -- uncropping ($69.1\%$), object swap ($74.5\%$), and re-texturing ($73.2\%$), demonstrating a consistent advantage over competing approaches.

\subsection{Ablation Analysis} \label{sec: experiments: ablation}

\begin{figure}
    \centering
    \vspace{-.5em}
    \includegraphics[width=\linewidth]{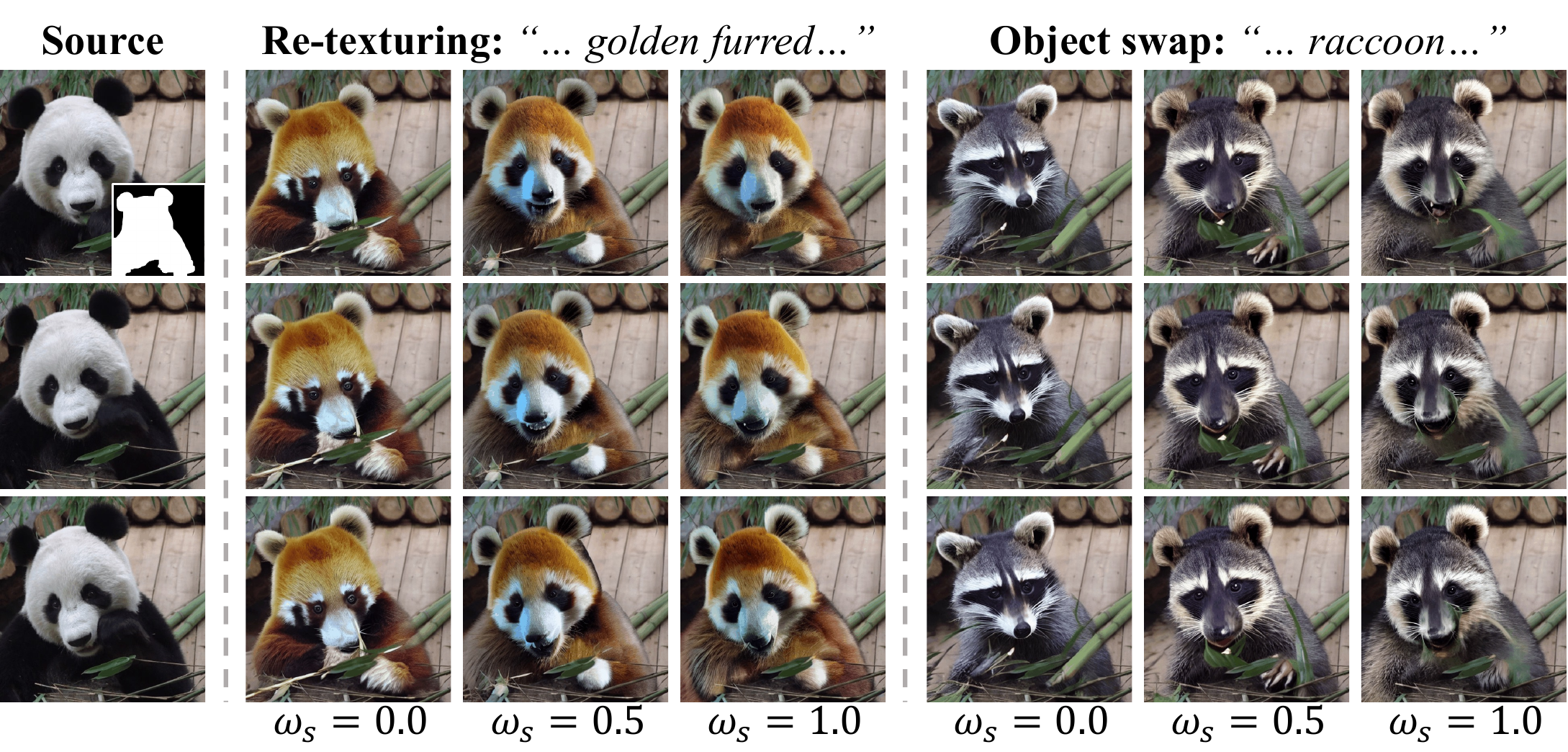}
    
    \caption{\textbf{Analysis of structure guidance.}
    We choose three evenly spaced frames from a video to highlight the impact of our structure guidance.
    To optimize results, the scale of our structure guidance must be tailored to each specific editing sub-task.
    }
    \vspace{-1.5em}
    \label{fig: ab_control}
    
\end{figure}

\noindent \textbf{Effect of structure guidance.}
In~\cref{fig: ab_control}, we exhibit the effects of varying the structure guidance scale, $\omega_s$, during the editing of a video of $16$ frames across different tasks. 
We highlight the first, middle, and last frames to demonstrate how structure guidance impacts the editing outcomes.
For the re-texturing task, where the objective is to transform the color of the panda's fur to golden, a higher $\omega_s$ ensures that more motion and structural details from the original video are retained. 
Conversely, with $\omega_s=0$, the motion of the generated golden-furred panda bears no correlation to the original video.
In the object swap, the goal is to replace the panda with a raccoon. 
Yet, when a high structure guidance scale is employed, the generated raccoon retains the shape of the panda, leading to a mismatch with the textual instruction.
In essence, the appropriate scale for structure guidance should be judiciously chosen based on the specific inpainting type the model is addressing.

\begin{figure}
    \centering
    \vspace{-1em}
    \includegraphics[width=\linewidth]{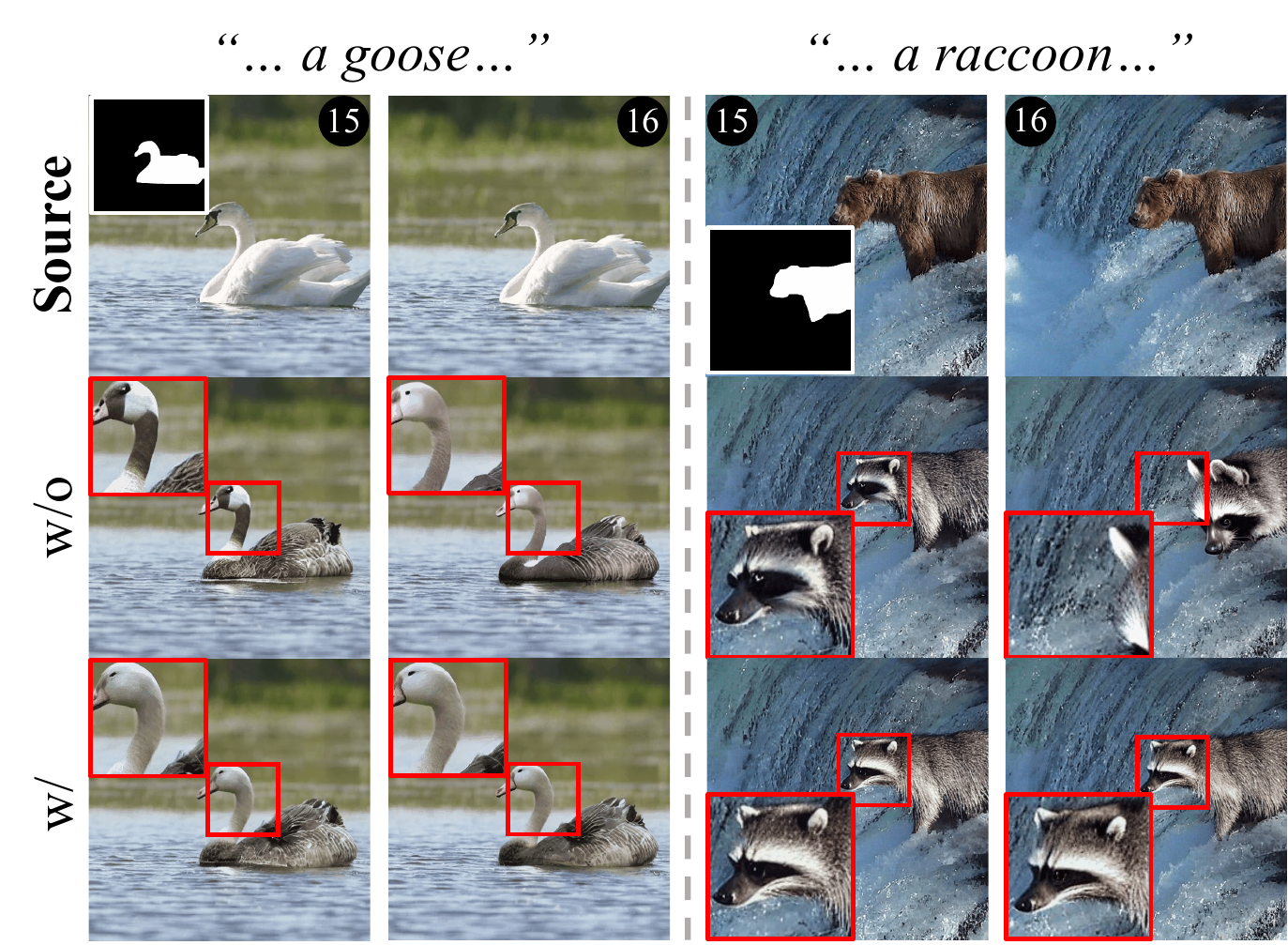}
    \vspace{-1.em}
    \caption{\textbf{Analysis of Temporal MultiDiffusion.}
    While direct sampling of two video clips (w/o) results in inconsistent content within the target area, temporal multi-diffusion sampling (w/) ensures the synthesis of longer videos with seamless transitions. 
    We highlight part of the target region in each video to better illustrate the effectiveness of our approach.
    }
    \label{fig: ab_multi}
    \vspace{-1.5em}
\end{figure}

\noindent \textbf{Temporal MultiDiffusion.}
We conduct longer video experiments on $32$-frame $6$-FPS videos, while our base model is trained on video clips consisting of $16$ frames.
As in~\ref{sec: methods: inference}, we can directly apply our model to the longer video due to the flexibility of our motion modules, but it will result in drastic quality degradation
Another naive approach is to apply our model twice on each non-overlapped 16 frames.
However, this approach significantly changes the identity of the generated content within the target region between the two segments. 
In ~\cref{fig: ab_multi}, we demonstrate the advantages of the Temporal MultiDiffusion pipeline we propose. 
We set attention guidance $\omega=0$ in this experiment to better illustrate the effect of Temporal MultiDiffusion sampling.
For illustration, we highlight the $15^{\text{th}}$ frame (the end of the first segment) and the $16^{\text{th}}$ frame (the beginning of the second segment).
As observed, there is a sudden change between the segments using the naive approach, \eg, there is a sudden shift in the texture pattern on the generated goose's neck and the position of the generated raccoon.
In comparison, the content transitions smoothly between any two consecutive frames, with the Temporal MultiDiffusion module.

\begin{figure}
    \centering
    \vspace{-1em}
    \includegraphics[width=\linewidth]{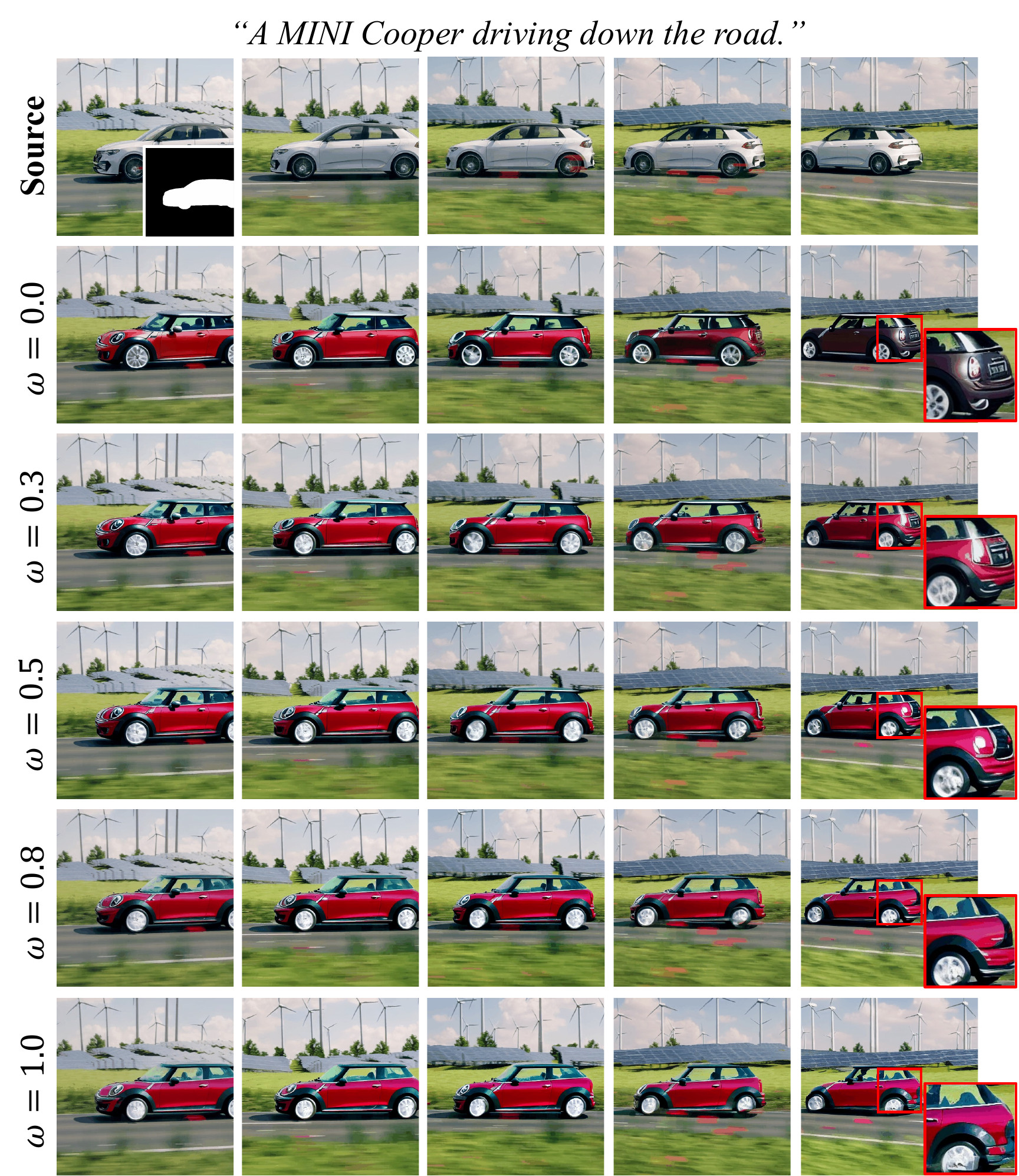}
    \vspace{-1.em}
    \caption{\textbf{Analysis of middle-frame attention guidance.}
    Adjusting $\omega$ yields editing results of differing quality. 
    }
    \label{fig: ab_attn}
    \vspace{-1.em}
\end{figure}

\noindent \textbf{Effect of middle-frame attention guidance.}
We delve into the effect of the attention guidance weight, denoted as $\omega$, as illustrated in \cref{fig: ab_attn}.
Our experimental videos are with $32$ frames and $6$ FPS.
Setting $\omega=0$ bypasses our proposed self-attention guidance mechanism. 
Although Temporal MultiDiffusion sampling ensures a smoother transition of content within the generated area, a noticeable identity shift persists throughout the video. 
For instance, with $\omega=0$, the rendered MINI Cooper gradually transitions from red to dark red. 
Our proposed attention guidance significantly mitigates this issue, ensuring a consistent identity from the first frame to the last. 
Conversely, setting a too large value for $\omega=1$ may introduce artifacts, as the keys and values in each self-attention layer \cref{equ: attn} are dominated by the middle frame signal.

\noindent \textbf{Discussion.}
Our model performance is limited by the learned motion module quality.
There are scenarios in which our model cannot generate content well especially when the editing prompt involves complicated actions, \eg, ``transforming the head of a horse from left to right" (in \emph{supplementary material}).
We believe a stronger motion module or a better text-to-video foundation model could further help improve the inpainting performance.
Another promising direction for future exploration is to turn the hyper-parameter of the structure guidance scale into a learnable parameter, controlled by the editing prompt.

\section{Conclusion}
\label{sec: conclusion}
We present \methodAbbr, a novel approach to address text-guided video inpainting.
Our model incorporates motion modules for better temporal consistency, and a structure guidance module for better structural fidelity to the original video.
At the sampling phase, we introduce a zero-shot inference pipeline, enabling our model to deal with videos with extended lengths.
Additionally, we propose a simple middle-frame attention guidance mechanism, greatly improving the identity consistency across video frames. 
Rigorous experiments show the effectiveness and robustness of \methodAbbr.

\noindent \textbf{Acknowledgments:}
This research has been partially funded by grants to D. Metaxas through NSF: 2310966, 2235405, 2212301, 2003874, and FA9550-23-1-0417.

\clearpage

{
    \small
    \bibliographystyle{ieeenat_fullname}
    \bibliography{main}
}

\clearpage
\appendix
\setcounter{section}{0}
\setcounter{figure}{0}
\setcounter{table}{0}
\renewcommand{\thefigure}{A\arabic{figure}}
\renewcommand{\thetable}{A\arabic{table}}

\twocolumn[{
\renewcommand\twocolumn[1][]{#1}
\maketitlesupplementary
\begin{center}
    \centering
    \includegraphics[width=1\linewidth]{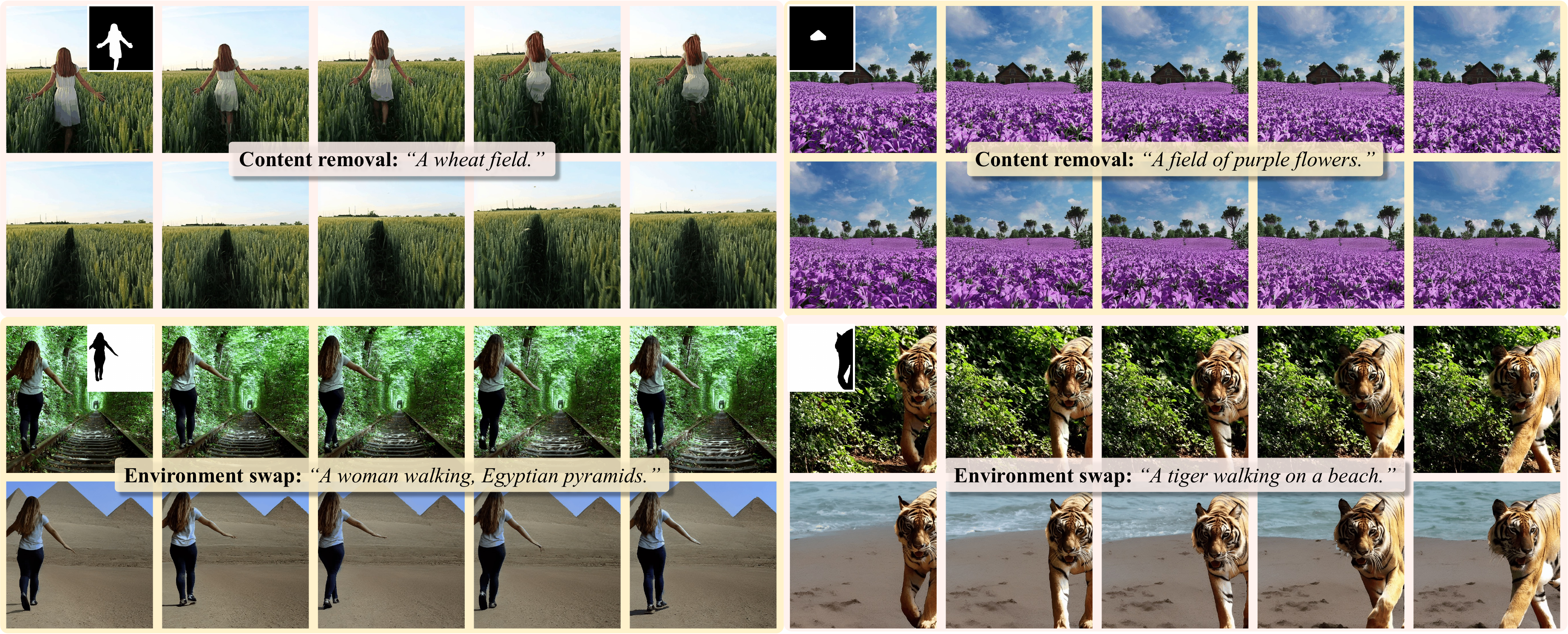}
    \vspace*{-1em}
    \captionof{figure}{
    \textbf{Other applications.} 
    We show how our approach can be applied to other video inpainting tasks, such as content removal and environment swap.
    }
    \label{fig: supp: applications}
\end{center}
}]

\section*{Overview}
The supplementary material accompanying this paper provides additional insights and elaborations on various aspects of our proposed method. 
The contents are organized as follows:

\begin{itemize}
    \item \textbf{Qualitative Results:} We showcase a broader range of qualitative results demonstrating the efficacy of every video inpainting type of \methodAbbr~ on videos of variable time duration. The results can be found in ~\cref{supp: sec: qualtative} of the supplementary material.
    
    \item \textbf{Application to Other Tasks:} ~\cref{supp: sec: other_tasks} presents the application of \methodAbbr~ on other text-guided video inpainting types.
    
    \item \textbf{Test-Time Efficiency Analysis:} An in-depth analysis of the test-time efficiency of our method is provided in ~\cref{supp: sec: efficiency}.
    
    \item \textbf{More Comparative Analysis:} Additional comparative studies are detailed in ~\cref{supp: sec: comp}.
    
    \item \textbf{Ablation Study:} We extend the ablation analysis mentioned in the main paper (\cref{supp: sec: ab}). 
    
    \item \textbf{Limitations:} ~\cref{supp: sec: limitations} is dedicated to discussing the limitations and potential areas for improvement in our method.
    
    \item \textbf{Extension to Text-to-Video Generation:} We explore the application of our proposed sampling pipeline to the domain of any-length text-to-video generation. The results of this exploration are presented in ~\cref{supp: sec: extension}.
\end{itemize}

For a more immersive experience, we encourage readers to look at the results in video format, available \href{https://zhang-zx.github.io/AVID/supp/index.html}{here}.

\section{Qualitative Results}
\label{supp: sec: qualtative}

In this section, we present an extensive collection of qualitative results that demonstrate the capabilities of our proposed method, \methodAbbr. 
This includes both the examples showcased in the main paper and additional results, offering a comprehensive view of our method's performance in various scenarios. 

To facilitate a more interactive and illustrative experience, these qualitative results are provided in video format. 
Readers are recommended to check these results in the first section of our \href{https://zhang-zx.github.io/AVID/supp/index.html}{accompanying webpage}. 
This visualization provides a more nuanced understanding of the temporal and visual qualities of our video inpainting results, as well as a deeper insight into the effectiveness of \methodAbbr~ in practical applications.

\section{Exploring Additional Inpainting Tasks}
\label{supp: sec: other_tasks}

This section delves into the adaptability of our \methodAbbr~ method to a broader spectrum of video inpainting applications, specifically focusing on content removal and environment swapping. 
Our experiments illustrate the versatility and effectiveness of \methodAbbr~ in handling diverse inpainting scenarios.

The experiments in this section are conducted using videos with $N^\prime = 24$ frames, corresponding to a duration of $4$ seconds. 
We set $\omega_s = 0.0$ in these experiments, meaning no structure guidance is applied. 
The results are visually represented in ~\cref{fig: supp: applications} of the supplementary material.

\subsection{Content Removal}
Video inpainting has been narrowly defined as content removal in previous literature~\cite{kim2019deep, xu2019deep, zeng2020learning}.
However, with diffusion models, we can enable multiple new inpainting tasks as introduced, which traditional approaches cannot handle. 
This work focuses mainly on content generation/editing guided by a given prompt and mask.
Nevertheless, our model does also support ``content removal". 

\begin{figure*}
\setlength{\linewidth}{\textwidth}
\setlength{\hsize}{\textwidth}
\centering
\includegraphics[width=1\linewidth]{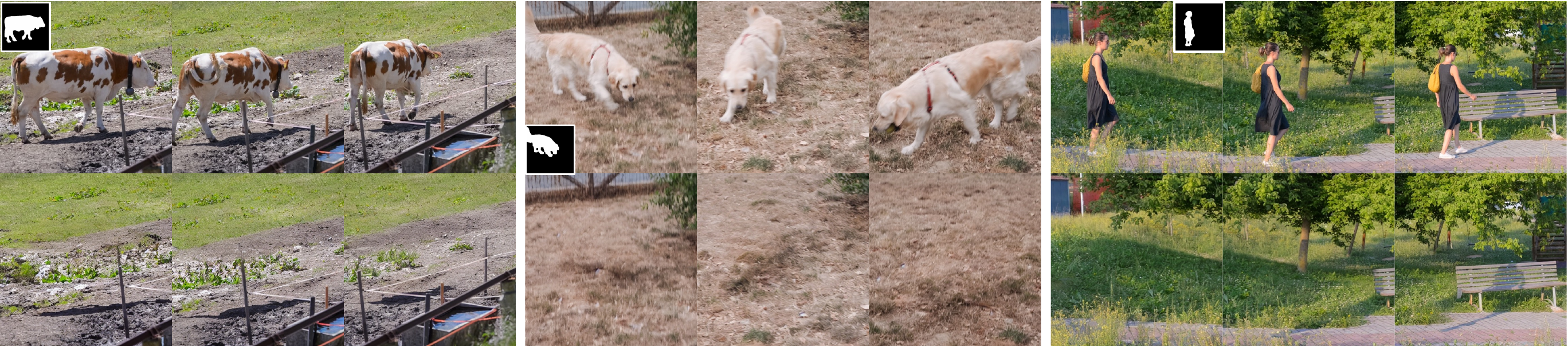}
\caption{\textbf{Content removal on DAVIS~\cite{Perazzi2016} dataset.} 
We apply our method for content removal on different videos in the DAVIS~\cite{Perazzi2016} dataset.
All frames of each video are passed to our model.
Frames shown in the figure are evenly distributed in each video.
We use prompts ``a field'', ``a grassland'', and ``a park'' respectively for these videos.
}
\label{fig: supp: davis}

\end{figure*}

The primary goal in content removal is to eliminate a specific object or element from the video while maintaining seamless integration with the surrounding content.
As demonstrated in the top block of \cref{fig: supp: applications}, our method initiates this process by generating a mask sequence targeting the object to be removed. 
Subsequently, we input a prompt such as ``A wheat field'' that describes the desired background, omitting any mention of the target object.
This strategy enables our model to effectively remove the object, replacing it with contextually coherent content that blends seamlessly with the surrounding area.
We further qualitatively evaluate our model on the popular DAVIS~\cite{Perazzi2016} dataset to better illustrate the ability of our method, as shown in ~\cref{fig: supp: davis}.

\subsection{Environment Swap}
The environment swap task involves altering the background or surrounding environment of a subject in the video.
Our method showcases its capability in environment swapping in the bottom block of \cref{fig: supp: applications}. 
By selecting the complement of the target region as the editing area, we can effectively modify the video's background. 
Through prompts describing the new environment, such as ``Egyptian pyramids'', our model can adeptly transform the surrounding setting, demonstrating its robustness in adapting to various inpainting contexts.

\subsection{Multiple Regions Inpainting}
\begin{figure}
    \centering
    \includegraphics[width=\linewidth]{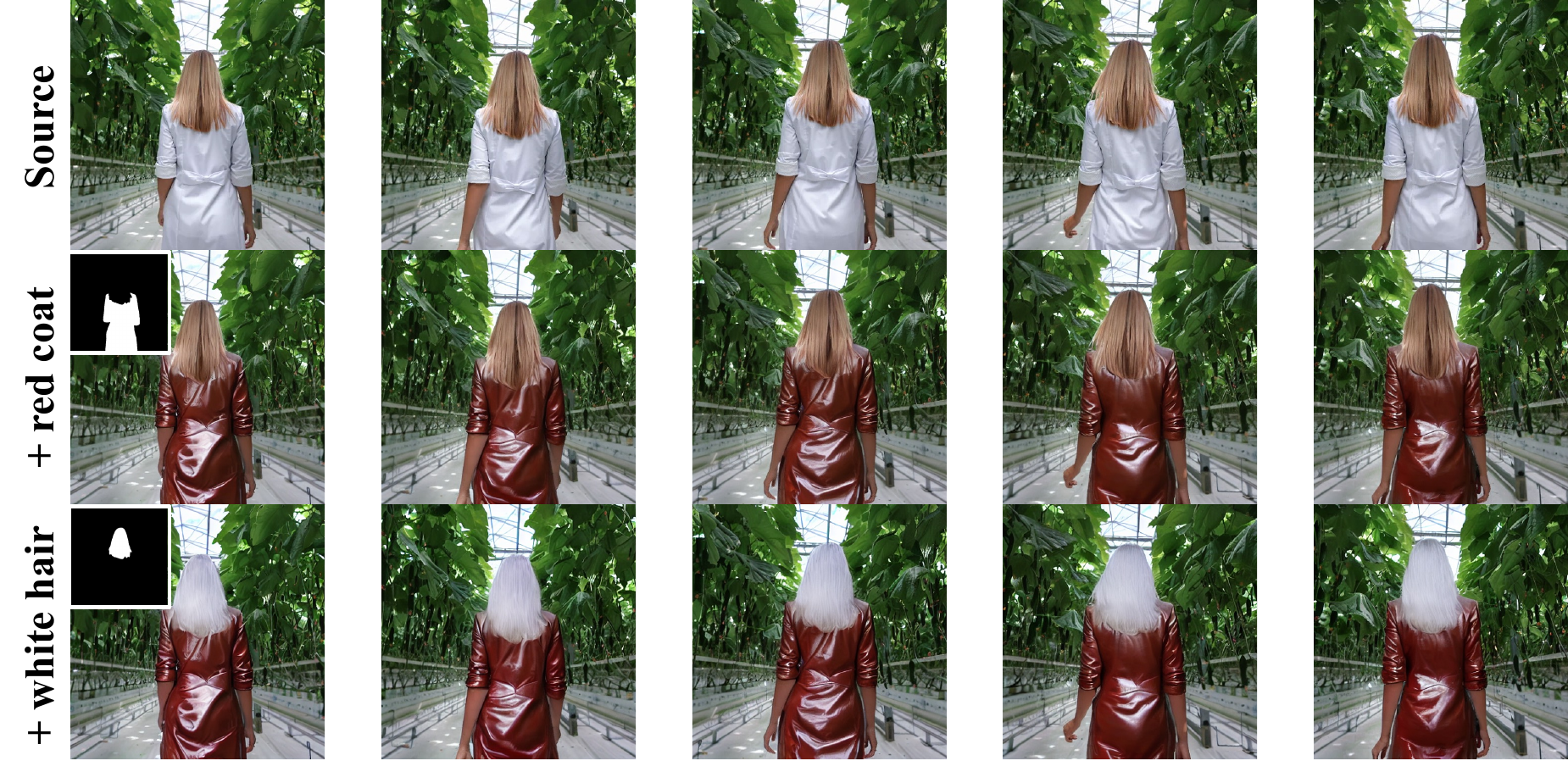}
    \caption{\textbf{Multiple objects inpainting.}
    We show how our approach can be applied to inpaint multiple objects in a video independently.
    }
    \label{fig: supp: multi_obj}
\end{figure}

Our method is not limited to inpainting one specific region in a video.
Independent inpainting can be achieved sequentially for multiple objects.
As shown in \cref{fig: supp: multi_obj}, we conduct re-texturing on two different regions, \ie coat, and hair, further demonstrating the effectiveness of our method in real-world applications.

\section{Test-time Efficiency}
\label{supp: sec: efficiency}

In this section, we extend the analysis to evaluate the test-time efficiency of our proposed Temporal MultiDiffusion pipeline. 
For simplicity, we bypass the structure guidance in this analysis. 
Building upon the foundation discussed in ~\cref{sec: methods: t2v inpainting}
of our main paper, our approach inflates an image inpainting diffusion model, inspired by AnimateDiff~\cite{guo2023animatediff}. 
This is achieved by transforming 2D layers into pseudo-3D format, allowing independent processing of each frame. 
To capture temporal correlations, we incorporate motion modules, realized through pixel-wise temporal self-attention.

Considering a video sequence with $N^\prime$ frames, a direct inference approach using all $N^\prime$ frames simultaneously leads to a temporal complexity of $\mathcal{O}({N^\prime}^2)$. 
The spatial complexity for attention layers, both self and cross attention, is $\mathcal{O}((HW)^2)$, with $H$ and $W$ being the spatial dimensions. 
Thus, our base model exhibits a computational complexity of $\mathcal{O}((HW)^2 \times {N^\prime}^2)$. 

The Temporal MultiDiffusion pipeline, however, segments the video into $n$ parts, each comprising $N$ frames, where $n=\lceil \frac{N^\prime - N}{\mathbf{o}} \rceil + 1$ and $\mathbf{o}$ represents the stride. 
This segmentation allows for independent calculation of each segment at every denoising step, reducing the temporal complexity to $\mathcal{O}(N^2 \times n)$.
With the incorporation of our middle-frame attention guidance mechanism, the spatial self-attention calculation effectively doubles, leading to a total temporal complexity of $\mathcal{O}(2 (HW)^2 \times( N^2 \times n))$. 
Notably, when $N^\prime >> N$, the complexity of our approach approximates to $\mathcal{O}((HW)^2 \times N^\prime)$, significantly more efficient than direct inference with $N^\prime$ frames. 
Additionally, our pipeline necessitates the calculation of only the segment containing the middle frame for initial attention guidance, while other segments can be processed in parallel, leveraging multi-GPU setups to expedite the process and mitigate potential GPU memory overflows in practical applications.

\section{More Comparison.}
\label{supp: sec: comp}

In this section, we engage in comparative experiments with TokenFlow~\cite{geyer2023tokenflow}, a state-of-the-art video editing method, to demonstrate the effectiveness of our proposed approach. 
We are following the methodology outlined in ~\cref{sec: experiments: comp} in our main paper, we assess the performance of TokenFlow against our method, particularly focusing on tasks such as re-texturing and object swapping.
Our evaluation utilizes the same set of videos and automatic metrics detailed in ~\cref{sec: experiments: comp}. 
This comparative study aims to provide an objective and quantifiable measure of each method's capabilities.

Despite TokenFlow's advanced editing capabilities, our experiments reveal a significant shortfall in its background preservation ability. 
Specifically, in the context of object swapping, TokenFlow scores $93.3$ compared to our method's $41.1$. 
Similarly, in re-texturing tasks, TokenFlow scores $90.8$ versus our $40.7$. 
This disparity can be attributed to TokenFlow's reliance on language-based guidance for determining the editing region, rather than using an explicit mask sequence. 
This approach undermines the method's suitability for precise video inpainting tasks, where maintaining contextual consistency is paramount.

An additional consideration is TokenFlow's use of DDIM inversion~\cite{song2020denoising} for temporal consistent latent initialization. 
In contrast, our method employs initialization from a standard Gaussian distribution. 
This fundamental difference in initialization strategy highlights TokenFlow's limitations in tasks where no guidance can be obtained from the source video in the target region, such as video uncropping. 
\begingroup
\setlength{\tabcolsep}{3.1pt} %
\renewcommand{\arraystretch}{0.9} %
\begin{table}[!htb]\centering
\small
\begin{tabular}{l ccc ccc}
Task &  \multicolumn{3}{c}{Object swap} & \multicolumn{3}{c}{Re-texturing$^*$} \\
\toprule
Metric  & 
        {\footnotesize BP} & {\footnotesize TA} & {\footnotesize TC} & 
        {\footnotesize BP} & {\footnotesize TA} & {\footnotesize TC} \\
\midrule
TF & 93.3 & 31.5 & {\bf 97.5}& 90.8 & {\bf 32.2} & {\bf 97.8 }\\
Ours & {\bf 41.1}& {\bf31.5 }& 96.5& {\bf 40.7}& 32.0& 96.3 \\
    \bottomrule
\end{tabular}
\caption{\textbf{Quantitative results.} We compare our method against TokenFlow (TF) ~\cite{geyer2023tokenflow} on different video generative fill sub-tasks and evaluate generated results using different metrics, including background preservation (BP $\times 10^{-3}$, $\downarrow$ better), text-video alignment (TA, $\uparrow$ better), and temporal consistency (TC, $\uparrow$ better). $^*$ indicates structure guidance is applied for our approach.
}
\label{table: quan_supp}

\end{table}
\endgroup

\section{More Ablation Analysis}
\label{supp: sec: ab}

\subsection{Temporal MultiDiffusion}
\label{supp: ablation: multidiff}

\begin{figure}
    \centering
    \includegraphics[width=1\linewidth]{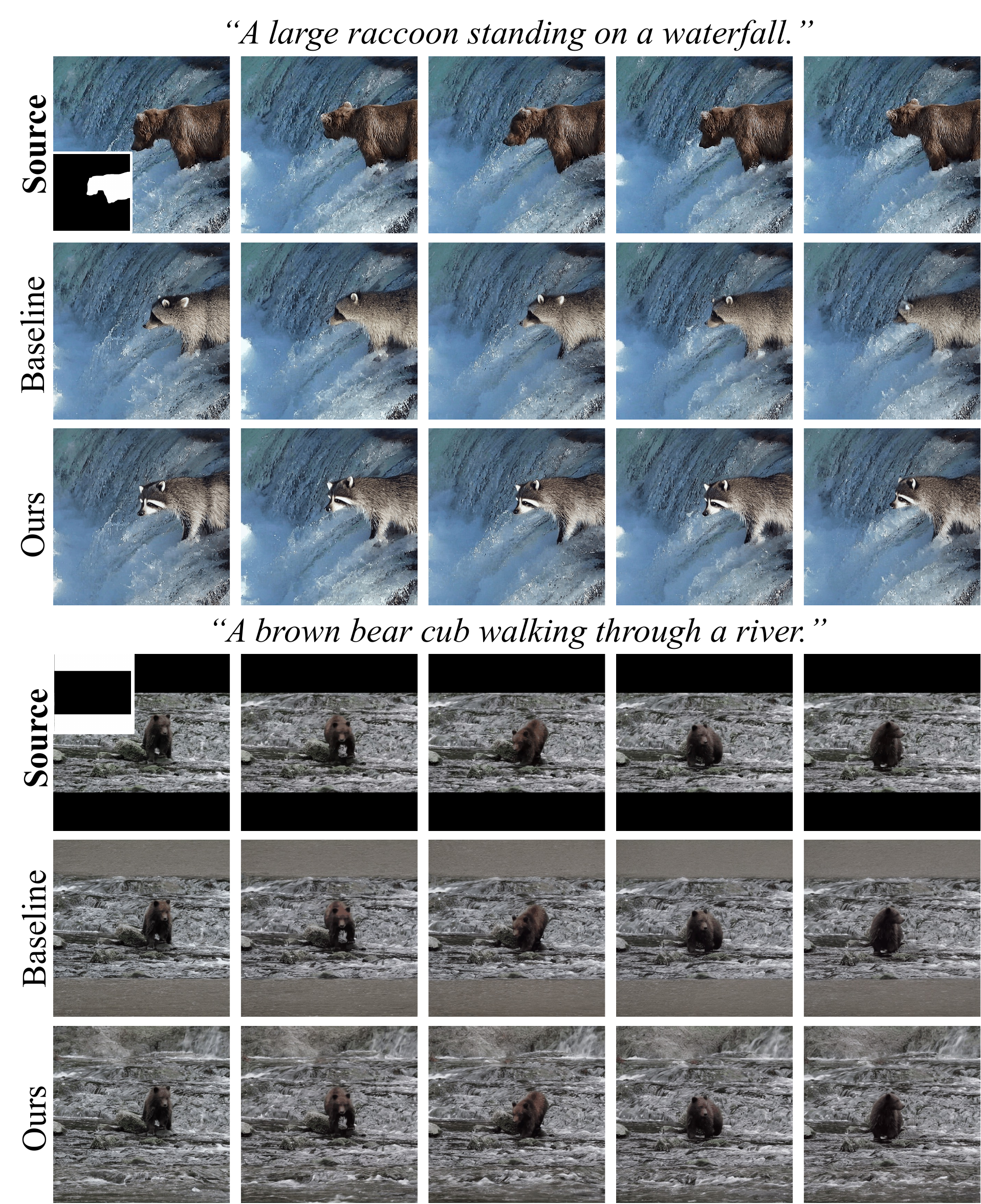}
    \caption{\textbf{Ablation analysis of temporal multi-diffusion.} 
        When we directly apply our model to video generative fill tasks of longer durations (specifically, 24 frames), it does not produce out-of-distribution results (row 2 and row 5). 
        However, there's a noticeable decline in the quality of the filled content when the length of the inference video differs from the training setup, the ear of the generated raccoon in the first case (row 2). 
        In the second case (row 5), the model fails to fill-in the target region with content that can seamlessly blend in with the rest area.
        In contrast, our method (row 3 and row 6) effectively addresses this issue, synthesizing high-quality content even for extended-duration videos.
        }
        \label{fig: supp: ablation: multi_diff}
\end{figure}

This section aims to evaluate the efficacy of our Temporal Multi-Diffusion sampling pipeline, especially in handling videos of varying durations. 
As discussed in ~\cref{sec: methods: inference} of the main paper, our model, while versatile, faces challenges in maintaining quality when dealing with frame counts different from those used in training. 
We address this issue by comparing the performance of our model using the Temporal Multi-Diffusion pipeline against its direct application on videos of different lengths.

Following the framework of AnimateDiff~\cite{guo2023animatediff}, our model incorporates sinusoidal position encoding~\cite{vaswani2017attention} within each temporal self-attention motion module. 
This encoding is pivotal in making the network aware of the temporal positioning of frames within a video clip. 
During training, we set the maximum length of this encoding to $24$ frames.

For our comparative analysis, we standardized the video length to $24$ frames. 
This approach allows for a balanced evaluation of our method against the baseline model. 
Notably, in these tests, we disabled the middle-frame attention guidance to ensure fairness in comparison.

As depicted in ~\cref{fig: supp: ablation: multi_diff}, we observed that direct inference with $24$ frames resulted in a significant decline in generation quality. 
In stark contrast, the adoption of our Temporal MultiDiffusion pipeline markedly improved performance.
This pipeline effectively preserved the model's generative quality, showcasing its robustness and adaptability to different video durations without compromising the visual fidelity of the generated content.

\subsection{Middle-frame Attention Guidance}
\label{supp: ablation: attn}

\begin{figure}
    \centering
    \includegraphics[width=1\linewidth]{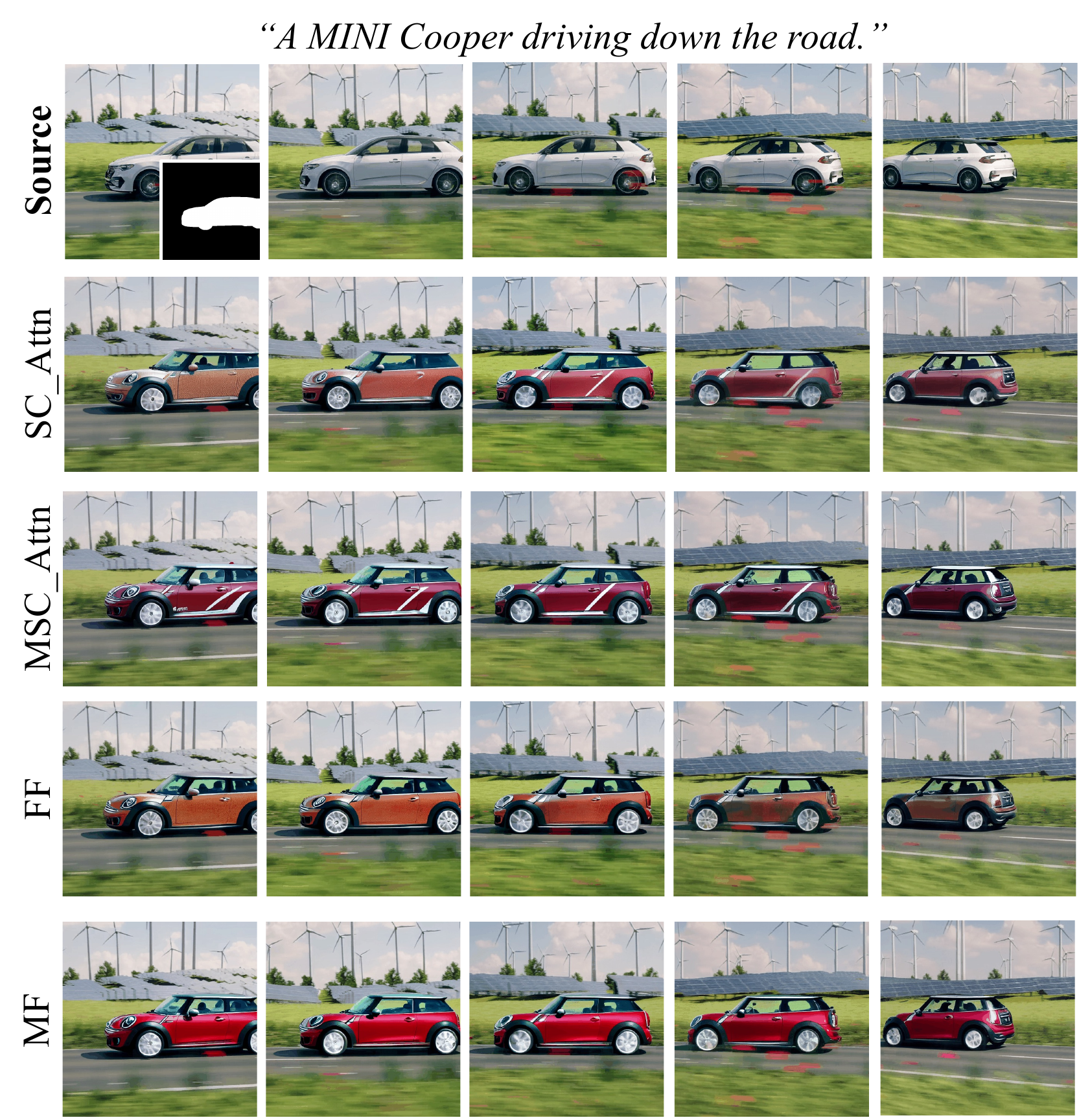}
    \caption{\textbf{Ablation analysis of attention guidance.} 
    We compare our middle-frame attention guidance approach (MF) with other temporal correlation modeling method variants, including Sparse-Casual Attention (SC\_Attn), Middle-frame Sparse-Casual Attention (MSC\_Attn), and First-frame attention guidance (FF).
        }
        \label{fig: supp: ablation: attn}
\end{figure}

In this section, we conduct an ablation study to underscore the efficacy of the middle-frame attention guidance mechanism introduced in our method.
This study is pivotal in demonstrating how our approach enhances temporal coherence in video inpainting tasks, a challenge extensively explored in recent works~\cite{wu2023tune, wang2023zero}.

\noindent \textbf{Attention mechanism:} 
Tune-A-Video~\cite{wu2023tune} proposes the use of Sparse-Casual Attention (SC\_Attn), which calculates the attention matrix between the current frame $\psi^i$ and two previous frames ($\psi^1$ and $\psi^{i - 1}$), as described in the following equation:
\begin{equation}
    \small
    \operatorname{Attention}(\psi^i) = \operatorname{softmax}\left(\frac{Q^i {K^i}^T}{\sqrt{d}} \right)V^i,
\end{equation}
where $Q^i = W^Q \psi^i$, $K^i = W^K \left[\psi^1,  \psi^{i - 1}\right]$, and $V^i = W^V \left[ \psi^1,  \psi^{i - 1}\right]$. 
A similar technique is also adopted in Pix2Video~\cite{ceylan2023pix2video}.
We adapt Sparse-Casual Attention within each segment of our Temporal MultiDiffusion pipeline.

SC\_Attn can be further extended to Middle-frame Sparse-Casual Attention (MSC\_Attn) by changing the anchor frame from the first frame within each segment to the middle frame in the whole video, $\psi^{\lceil N^\prime / 2 \rceil}$.

\noindent \textbf{Key frame selection:}
Additionally, we experiment with using the first frame of the video as the guidance frame, modifying our self-attention computation as per Equ. 6 in the main paper:
\begin{equation}
\small
\label{equ: attn_ff}
\begin{split}
    \small
    \operatorname{Attention}(\psi^i) =& \operatorname{softmax}\left(\frac{Q^i {K^i}^T}{\sqrt{d}} \right)V^i \cdot (1 - \omega)+ \\
                                  & \operatorname{softmax}\left(\frac{Q^i {K^{1}}^T}{\sqrt{d}}  \right )V^{1} \cdot \omega.
\end{split}
\end{equation}
We employ an attention guidance weight of $\omega=0.3$ for this variant.

\noindent \textbf{Results and discussion:}
Our experiments, visualized in ~\cref{fig: supp: ablation: attn}, demonstrate the varying degrees of success in addressing identity shift issues. 
The Sparse-Casual Attention (row 2) struggles to prevent identity shifts due to using different key-frames within each segment. 
Middle-frame Sparse-Casual Attention (row 3) yields better identity preservation, although inconsistencies in the generated patterns can still be observed. 
The approach using the first frame as guidance (row 4), while maintaining pattern stability, still exhibits significant color variance between the first and last frames.

In contrast, our proposed middle-frame attention guidance mechanism (row 5) excelled in preserving both the color and pattern on the car consistently throughout the video. 
This result not only highlights the superiority of our method in maintaining temporal coherence but also emphasizes the critical role of strategic frame selection in attention guidance mechanisms for video inpainting tasks.

\subsection{Test-time Masks Accuracy}
\label{supp: ablation: mask}
\begin{figure}
    \centering
    \includegraphics[width=1\linewidth]{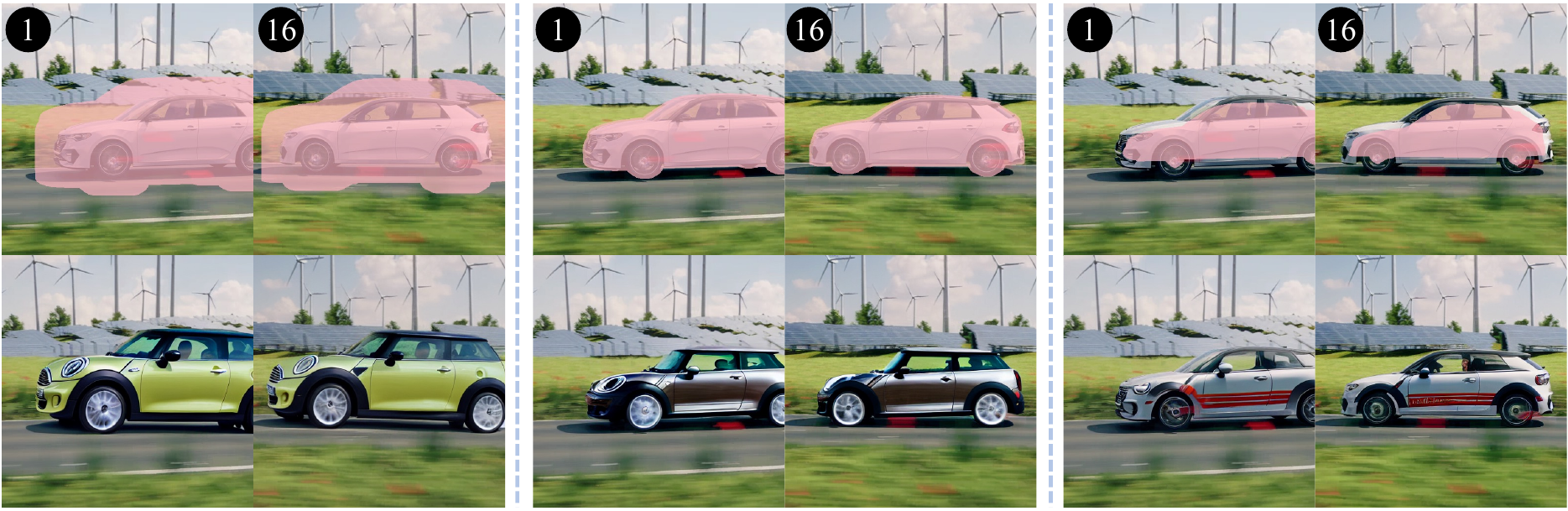}
    \caption{\textbf{Ablation analysis of mask accuracy.} 
    We explore the robustness of our method using different mask regions.
    Here we show $3$ examples using ``inaccurately" expanded, accurate, and ``inaccurately" eroded masks with the same prompt of ``Mini Cooper".
        }
        \label{fig: supp: ablation: mask}
\end{figure}

Due to using random synthetic masks during training, our model is very robust to inaccurate masks.
As shown in \cref{fig: supp: ablation: mask}, our method can successfully inpaint the video following the given text prompt when the mask region is significantly larger than the region size.
However, when the mask area can not cover the whole to-be-replaced object, our method will fail to modify the shape of the object due to the preservation of out-of-region details.

\section{Limitations}
\label{supp: sec: limitations}
\begin{figure}
    \centering
    \includegraphics[width=\linewidth]{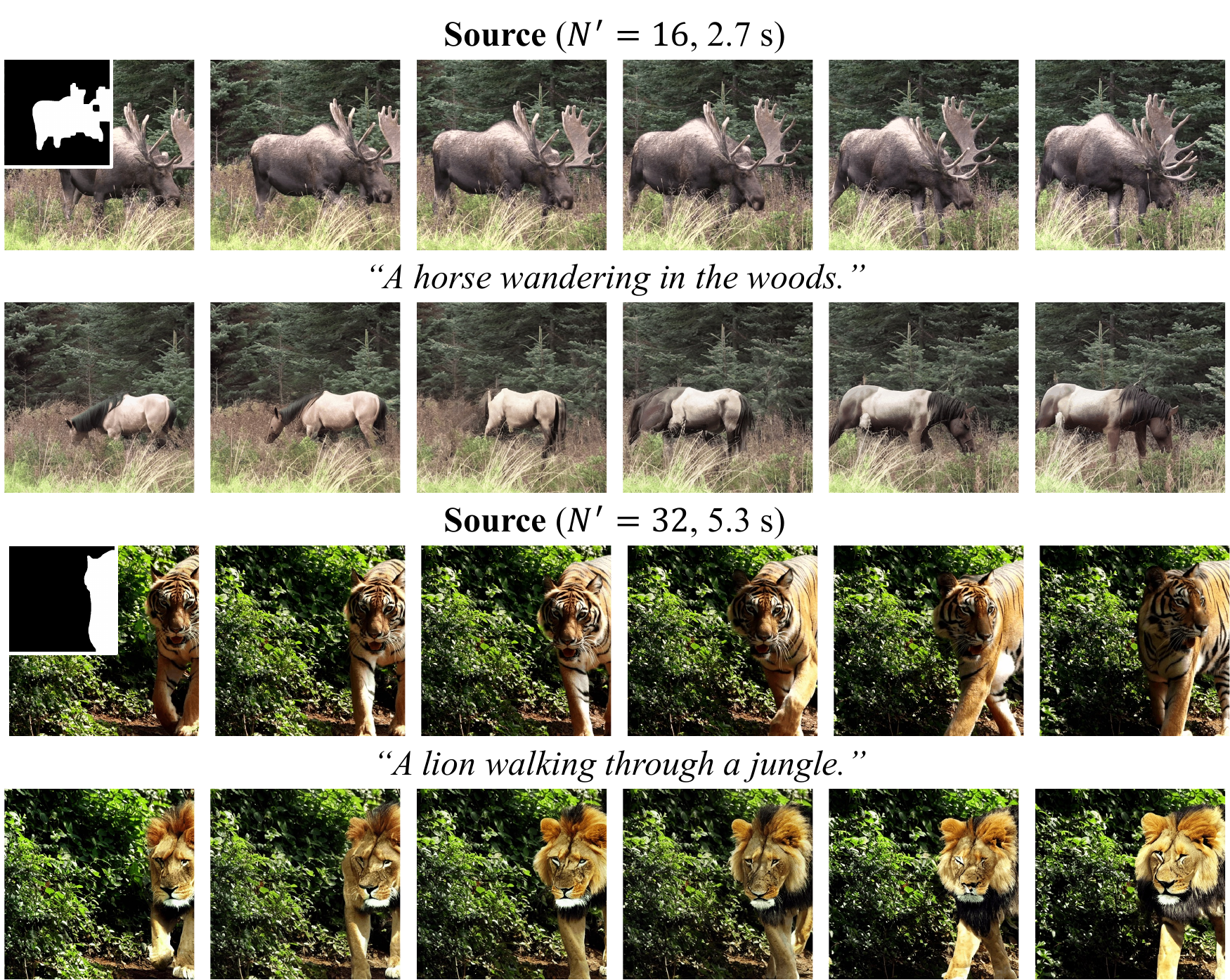}
    \caption{\textbf{Failure cases.}
    We showcase where our method fails to generate results with high fidelity.
    In the first case, the head of the horse first disappears and then reappears, while in the second case the left foot of the generated lion moving forward, it goes through the right foot of the lion.
    Please refer to the video results for a better illustration. 
    }
    \vspace{-1em}
    \label{fig: supp: limitations}
\end{figure}

In this section, we delve into specific instances where our method can not yield the desired results, as illustrated in \cref{fig: supp: limitations}. 
These failure cases, particularly in scenarios involving complex actions, offer crucial insights into the limitations of our current approach and highlight areas for future improvement.

As shown in the first case, in an attempt to generate a horse moving its head from left to right, our method fails to generate plausible results. 
Instead of showing a smooth head movement, the generated video exhibits the head of the horse disappearing and reappearing on the right side. 
Concurrently, the body of the horse undergoes an unnatural morphing, transitioning from facing left to right with only minor shape changes.
Another challenging scenario involves a lion walking forward. 
The generated video inaccurately shows the left foot of the lion moving through its right foot, an evident deviation from natural movement.
For both cases, we recommend viewing the video results for a more comprehensive understanding of these issues.

As noted in the main paper, these limitations are perhaps due to that our current foundation text-to-video model lacks high-quality motion generation capability.
We believe that enhancing the model with more advanced capabilities, especially in interpreting and rendering complex actions, can further improve the quality.
A stronger foundation model may also offer better comprehension of intricate movements and interactions, thereby producing more accurate and realistic video content.

Besides the limitations discussed above, we admit our model fails at handling discontinuity, especially objects moving out and back to the video.
Such an issue could be mitigated with a more deliberate cross-clip attention injection mechanism, which is a critical direction to further improve the robustness.

\section{Any-length Text-to-Video Generation}
\label{supp: sec: extension}

In this section, we explore the application of our proposed inference pipeline to existing text-to-video generation frameworks, such as AnimateDiff~\cite{guo2023animatediff}, demonstrating its potential in facilitating any-length text-to-video generation. 
This exploration serves as a testament to the versatility and adaptability of our method in broader video generation contexts.
We have included preliminary results of this extension on the \href{https://zhang-zx.github.io/AVID/supp/index.html}{accompanying webpage}. 

A promising direction for future research lies in the realm of sequential storytelling through video. 
This involves the idea of performing inference with a series of text prompts, effectively guiding the attention mechanism to evolve in tandem with the narrative. 
Such an approach could revolutionize how stories are visually narrated, aligning the generated video content with a progressive textual storyline.

\end{document}